\documentclass[hidelinks,onefignum,onetabnum]{siamart220329}

\usepackage{amsmath,amsfonts,bm}

\def\eqref#1{equation~\ref{#1}}

\def\1{\bm{1}}

\DeclareMathAlphabet{\mathsfit}{\encodingdefault}{\sfdefault}{m}{sl}
\SetMathAlphabet{\mathsfit}{bold}{\encodingdefault}{\sfdefault}{bx}{n}

\usepackage{lipsum}
\usepackage{amsfonts}
\usepackage{graphicx}
\usepackage{epstopdf}
\usepackage{algorithmic}
\usepackage{siunitx}
\usepackage{subcaption}
\usepackage{multirow}
\usepackage{calc}
\usepackage[colorinlistoftodos]{todonotes}

\ifpdf
  \DeclareGraphicsExtensions{.eps,.pdf,.png,.jpg}
\else
  \DeclareGraphicsExtensions{.eps}
\fi

\newcommand{\norm}[1]{\left\lVert#1\right\rVert}

\newcommand{\bfB}{{\bf B}}

\newcommand{\bfH}{{\bf H}}
\newcommand{\bfI}{{\bf I}}
\newcommand{\bfJ}{{\bf J}}
\newcommand{\bfK}{{\bf K}}

\newcommand{\bfc}{{\bf c}}

\newcommand{\bfg}{{\bf g}}

\newcommand{\bfk}{{\bf k}}

\newcommand{\bfx}{{\bf x}}
\newcommand{\bfy}{ {\bf y}}

\newcommand{\bfq}{{\bf q}}

\newcommand{\bfr}{{\bf r}}
\newcommand{\bff}{{\bf f}}
\newcommand{\bfv}{{\bf v}}

\newcommand{\bfz}{{\bf z}}

\newcommand{\bftheta}{{\boldsymbol \theta}}

\newcommand{\bflambda}{{\boldsymbol \lambda}}

\newsiamremark{remark}{Remark}
\newsiamremark{hypothesis}{Hypothesis}
\crefname{hypothesis}{Hypothesis}{Hypotheses}
\newsiamthm{claim}{Claim}

\headers{Neural DAEs}{T. Boesen, E. Haber, and U. M. Ascher}

\title{Neural DAEs: Constrained neural networks\thanks{Submitted to the editors 21-05-2023.
}}

\author{Tue Boesen\thanks{Department of Earth Ocean and Atmospheric Sciences, University of British Columbia, Canada 
  (\email{tue.boesen@protonmail.com}, \email{ehaber@eoas.ubc.ca}).}
\and Eldad Haber\footnotemark[2]
\and Uri M. Ascher\thanks{Department of Computer Science, University of British Columbia, Canada 
  (\email{ascher@cs.ubc.ca}).}}

\usepackage{amsopn}

\ifpdf
\hypersetup{
  pdftitle={Neural DAEs: Constrained neural networks},
  pdfauthor={T. Boesen, E. Haber, and U. M. Ascher}
}
\fi

\begin{document}

\maketitle

\begin{abstract}
This article investigates the effect of explicitly adding auxiliary algebraic trajectory information to neural networks for dynamical systems. We draw inspiration from the field of differential-algebraic equations and differential equations on manifolds and implement related methods in residual neural networks, despite some fundamental scenario differences. Constraint or auxiliary information effects are incorporated through stabilization as well as projection methods, and we show when to use which method based on experiments involving simulations of multi-body pendulums and molecular dynamics scenarios. Several of our methods are easy to implement in existing code and have limited impact on training performance while giving significant boosts in terms of inference.
\end{abstract}



\section{Introduction}

Many scientific simulations {of dynamical systems} have natural invariants that can be expressed by constraints. Such constraints represent conservation of some quantities of the system under study. For example, in molecular dynamics, bond lengths between atoms are assumed fixed in time, in incompressible fluid flow and Maxwell's equations, the divergence of the velocity and magnetic fields vanishes at any point in space and time.
Such additional information about the dynamics can be crucial if we are to keep the simulations faithful. As a result, a wealth of techniques have been proposed to conduct simulations that honour the constraints at least approximately \cite{weiglhofer1994medium,ap,allen2004introduction}.

In recent years,  machine learning based techniques, and in particular deep neural networks, have been taking a growing role in modelling physical phenomena. In some cases, such techniques are used as inexpensive surrogates of the true physical dynamics and in others they are used to replace it altogether (see e.g. \cite{wang2018deepmd, degiacomi2019coupling, miyanawala2017efficient}).  
These techniques use the wealth of data, either observed or numerically generated, in order to ``learn'' the parameters in a neural network, so that the data are fit to some accuracy.
The network is then expected to perform well on new data, outside of the training set and yield simulation results that are almost as accurate and reliable as a physical simulation, at a significantly lower computational effort. 

From classical simulations, we know that it is often more important to accurately obey the additional constraint information than it is to satisfy the underlying ordinary/partial differential equation (ODE/PDE) system. 
For instance, a car crash simulation would be considered useless if the distance between the vehicle's front wheels varies in time before the crash, whereas some inaccuracy in the velocity would be more easily tolerated.
The hope in standard training procedures is that by fitting the data, the network will ``learn'' the algebraic constraints and embed them in the weights implicitly. This, however, has been demonstrated to be insufficient in many cases \cite{wah2001violation, saad2022guiding, hansen2023learning}.

As we show in this paper, on some simple examples, neural networks may be able to approximately learn the dynamics but solutions can stay off the constraints manifold. This is to be expected if the validation loss does not vanish, since it does not give preference to the invariants over other characteristics of the system.
This phenomenon leads to erroneous  results that violate simple underlying physical properties. The question then is: {\em How should additional constraints information be incorporated into a neural network architecture to best ensure that the constraints are honoured?}

The idea of adding constraint information to a network is essentially a continuation of the ongoing process of connecting mathematics with machine learning and explicitly adding known information into a neural network rather than having the network implicitly learn it~\cite{willard2021integrating}. Equivariant networks are one example of this, where the symmetry of a problem is explicitly built into the neural network \cite{thomas2018tensor}. 
Other work on adding constraints to a neural network includes~\cite{stewart2017label,xu2018semantic} which add an auxiliary regularization term to the loss function; 
Physics-informed neural networks (PINNs) is another closely related yet distinct branch of combining mathematics with machine learning. PINNs use neural networks to obtain solutions to already specified differential equations. PINNs gained popularity following the work by \cite{raissi2019physics}, and have been actively researched in the previous years \cite{rackauckas2020universal,krishnapriyan2021characterizing, lu2021physics,negiar2022learning,hansen2023learning,sridhar2023guaranteed}. Notably \cite{lu2021physics} adds hard constraints as opposed to the traditional soft constraints introduced via loss functions, while \cite{negiar2022learning} adds a PDE layer to a neural network, such that the PDE constraints are enforced as hard constraints on the interior domain. In both cases they show that hard constraints generally leads to superior solutions compared to soft constraints, which is similar to our findings. 
Our method is fundamentally different from PINNs, since our method requires no knowledge about the differential equation apart from the knowledge of the constraints it must fulfill. In this sense our work is closely related to neural ODEs \cite{haber2017stable,neuroODE}, which is another example of connecting mathematics with machine learning. Neural ODEs has inspired a variety of extensions, \cite{kidger2022neural} work on neural controlled differential equations and neural stochastic differential equations are two such extensions, while ours is another.

In this paper we introduce a neural network architecture that allow incorporation of additional information in a dynamical system. Motivated by the mathematical field of differential algebraic equations (DAE) and Neural ODEs, we study two different approaches to solve the problem.
The first is the (approximate) incorporation of the constraints into the network by Lagrange multipliers, and the second employs so-called stabilization techniques that aim to penalize outputs that grossly violate the constraints. 
Both approaches have similar counterparts in the physical simulation world, and in particular in systems of DAEs~\cite{ap}. 
Note, though, that the paradigm is not entirely the same, because the well-known effect of a drift off the constraint manifold (as seen in physical simulations where constraints are ignored) is automatically softened when learning from solution data (which implicitly, if inaccurately, take constraint information more directly into account).
Our methodology is designed for residual neural networks, however it can be used and adopted for other architectures as well.
We experiment with it on a number of well-known problems, 
focusing on molecular dynamics (MD) applications~\cite{allen2004introduction}, which enables us to incorporate a variety of algebraic constraints
often resulting in significant performance improvement.
 
The rest of this paper is organized as follows: \Cref{sec:prelim_theory} describes the neural ODE setting amended by additional algebraic information. \Cref{sec:theory} then presents various ways that constraint effects can be introduced in neural networks. \Cref{sec:model_problems} introduces our model problems, and describes relevant constraints for those problems, while \cref{sec:experiment} performs a series of experiments based on them using constrained neural networks. The paper is wrapped up in \Cref{sec:dc} with a discussion and conclusions.   


\section{Residual Networks and Constraints}
\label{sec:prelim_theory}

We consider the problem of statistical learning where data pairs $(\bfx_i, \bfy_i), i=1, \ldots n$, are given, 
and we assume that $\bfx \in {\cal X}$ and $\bfy \in {\cal Y}$, where ${\cal X}$ and ${\cal Y}$ are appropriate spaces. Our goal is to find a function ${\bff}$ depending on
parameters $\bftheta$ that satisfies
\begin{eqnarray}
\label{nn}
\bff(\bfx, \bftheta) = \bfy, \quad \bfx \in {\cal X}, \bfy \in {\cal Y}.
\end{eqnarray} 

We focus our attention on functions $\bff(\cdot,\cdot)$ that are mapped by neural networks. 
Such functions typically contain a sequence of multiplications by ``learnable'' matrices followed by nonlinearities. 
In this work we particularly focus 
on the continuous form of residual network architectures that reads
\begin{subequations}
\label{resnet}
\begin{align}
\label{resnet1}
\bfz_0 &= \bfK_o \bfx, \\
\dot \bfz &= \bfg(\bfz (\tau), \bftheta(\tau)), \quad \tau \in [0, 1], \label{resnet2} \\
\bfy_p  &= \bfK \bfz(1).
\label{resnet3}
\end{align}
\end{subequations}
The matrix $\bfK_o$ typically embeds input data $\bfx \in {\cal X}$ in a vector $\bfz$ in a larger space ${\cal Z}$, and $\bfz(0) = \bfz_0$. Next, the residual network discretizes the ODE \eqref{resnet2} (using for instance the forward Euler method) and uses a number of layers with learnable parameter $\bftheta$, each wrapped in a non-linear activation function $\bfg$. 
Finally, the larger space is closed by a learnable closing matrix $\bfK$, which gives us the model prediction $\bfy_p$ that in an ideal setting is equal to $\bfy$.

For the class of problems considered here, we assume that there is a given vector function $\bfc(\cdot)$ such that
\begin{equation}
\bfc(\bfy) = \mathbf{0}.
\label{eq:const_data}
\end{equation}
The ODE \cref{resnet2} with the initial condition \cref{resnet1} represent a trajectory, $\bfz(\tau)$, in the high dimensional space ${\cal Z}$ for each $\tau$. 
This trajectory can be projected into the low-dimensional physical space satisfying equation~\cref{resnet3} by requiring

\begin{equation}
\bfc (\bfK \bfz (\tau) ) = \mathbf{0} , \quad 0 \leq \tau \leq 1. 
\label{const}
\end{equation}


In reality, a trained neural network is an inexact model that only approximates the solution. 
Therefore, in general, $\bfc(\bfK \bfz)$ is not automatically close to $\mathbf{0}$ and this can yield results that are physically infeasible even for models that otherwise provide reasonable results, as we later demonstrate in \cref{fig:npendulum_final_snapshots}.
Our goal is to modify the architecture given in~\cref{resnet} such that the additional information given by~\cref{const} is addressed. 
We next discuss four such approaches that can be used.

\section{Adding auxiliary constraint information} 
\label{sec:theory}
In this section, we discuss four methods towards better honoring the additional information given in~\eqref{const}. The following four subsections introduce these four methods, which we call: ``auxiliary loss'', ``end constraints'', ``smooth constraints'' and ``penalty'', respectively. 

Note that, strictly speaking, the constraint \eqref{const} is important and valid only at $\tau = 1$. 
The first two methods, discussed in sections~\ref{sec:regularization} and \ref{sec:end_projection} therefore focus on modifications at the right end of the neural time interval (the last layer of the neural network).
``auxiliary loss'' does this by adding a regularization term to the loss function, while ``end constraints'' does this by projecting the last layer output onto the constraint.

The last two methods, described in sections~\ref{sec:smooth_projection} and \ref{sec:stabilization}, seeks to honour \eqref{const} on the entire interval in $\tau$, which means the constraint is introduced smoothly through the layers of the neural network.
The method ``smooth constraints'' does this by projecting the output of each layer of the neural network onto the constraint, while the ``penalty'' method does it by adding an additional stabilization term to each layer of the neural network which seeks to push each layer towards honouring the constraint.

The ``auxiliary loss'' method is a well-known method, but the other three methods are all novel.

 
\subsection{Auxiliary regularization}
\label{sec:regularization}
The simplest method for incorporating constraint information is to add an auxiliary regularization term to the loss function. Let $\mathcal{L}$ be the traditional loss. 
We then define
\begin{equation}
    \mathcal{L}_{\eta} = \mathcal{L} + {\frac \eta2} \bfc(\bfy_p)^\top \bfc(\bfy_p),
\end{equation}
where the positive parameter $\eta$ determines the strength of the regularization.
This does not modify the architecture of the network used to honour the constraints, however, the idea is, that by adding such terms to the loss it is possible to control the amount of constraint violation at the cost of decreased data inference. However, this does not guarantee that $\bfc$ is actually small at inference, as also shown in \cite{saad2022guiding}.


\subsection{Projecting the final state onto the constraint manifold}
\label{sec:end_projection}

Another approach that has been used for problems in image processing where the output is bounded, is to project the final state onto the constraint manifold. 
To this end, the network is used as is, obtaining a vector $\bfy_p$. We then look for a perturbation $\delta \bfy_p$ such that 
\begin{equation*}
\bfc(\bfy_p + \delta \bfy_p) = \mathbf{0}.
\end{equation*}
This problem is not well-posed as there are many such perturbations. 
Therefore, we look for the perturbation with 
minimal norm, which results in the iteration 
\begin{eqnarray}
\label{eq:proj}
\bfy_{p,j+1} = \bfy_{p,j} - \bfB \bfJ(\bfy_{p,j})^{\top}\bfc(\bfy_{p,j}), \quad j = 0, 1, \ldots, M,
\end{eqnarray}
where $\bfy_{p,0}=\bfy_p$ is the output of the unconstrained residual neural network, $\bfJ = \nabla_{\bfy_p} \bfc(\bfy_p)$ is the Jacobian, and $\bfB$ is an approximation to $(\bfJ^{\top} \bfJ)^{-1}$.
 
Since the network in this method observes the constraint only at the end $\tau = 1$, 
solving for the projection may require many iterations. Nonetheless, it is important to note that there are no learnable weights in the projection. Thus, by using implicit differentiation it is possible to write an analytic backward function and therefore, there is no need to hold all the states when computing the projection in order to compute the derivative, which was also done in \cite{negiar2022learning}.
Using the constraint at the end can be thought of as a ``final layer'' before the results are sent into the loss function.
 
Projecting the state at the end has an advantage that it is simple and requires minimal changes to any existing code. However, it may have some serious drawbacks. In particular, the dynamics in \eqref{resnet} can lead to states that are very far from satisfying the constraint. In this case, the constraint can be difficult to fulfil and the training can be difficult due to very large changes in the last layer. 
We therefore explore two different techniques that can be applied in order to follow the constraint throughout the network, at least approximately.
 
 
\subsection{Projecting the state onto the constraint manifold throughout the network}
\label{sec:smooth_projection}
 
To change the architecture we recall that when constraints are added to an ODE one obtains a DAE. To this end another parameter vector function (a Lagrange multiplier) is added to the system 
and~\eqref{resnet2} is replaced with 
\begin{equation}
 \label{dae}
 \dot \bfz = \bfg(\bfz, \bftheta) + \bfK^\top \bfJ(\bfK \bfz)^{\top} \bflambda,  \quad \tau \in [0, 1], 
 \end{equation}
while requiring~\eqref{const} to hold.
(See~\cite{ap} for derivation.)
 
Consider first using an explicit stepping method to discretize the system with respect to $\bfz$ 
\begin{equation*}
\bfz_{k+1}^* = \bfz_k + h \delta \bfq(\bfz_k),
\end{equation*}
where $k$ refers to the $k$-th discrete layer in the neural network, $h>0$ is the discrete step length, and $\delta \bfq$ is a single discretization of the ODE with respect to $\bfz$ around $\bfz_k$. 
With forward Euler it has the form 
\begin{equation*}
\delta \bfq(\bfz_k) = \bfg(\bfz_k, \bftheta_k).
\end{equation*}
However, from classical simulations it is known that forward Euler may have poor stability when nearly imaginary eigenvalues are present, and similar results have also been found for neural networks \cite{krishnapriyan2023learning}. Implicit methods were also considered, but were discarded as they are generally considered too expensive and not needed when working with neural networks as discussed in \cite{kidger2022neural}. Because of this, we use the classical, explicit Runge-Kutta 4th order (RK4) discretization which can be written as \cite{ascher1998computer}. 
\begin{equation}
\delta \bfq_{RK4}(\bfz_k) = 
{\frac 16} \left(\bfk_1
+ 2 \bfk_2 
+ 2\bfk_3 
+ \bfk_4 
\right)
\label{eq:RK4}
\end{equation}
where each $\bfk_i = \bfk_i (\bfz_k, \bftheta_k)$ is the Runge-Kutta stage that requires a single application of a resnet. 

Next, to obtain $\bflambda = \bflambda_{k+1}$ we substitute 
\begin{equation}
\bfz_{k+1} =  \bfz_{k+1}^*  + h \bfK^\top \bfJ(\bfK  \bfz_{k+1}^*)^{\top} \bflambda
\label{eq:update}
\end{equation}
into the constraint \eqref{const}, obtaining
\begin{equation*}
\bfc(\bfK \bfz_{k+1}) = \bfc\left(\bfK (\bfz_{k+1}^*  + h \bfK^\top \bfJ^{\top} \bflambda)\right) = \mathbf{0},
\end{equation*}
with $\bfJ = \bfJ(\bfK \bfz^*_{k+1})$.
By linearizing at $\bfz^*_{k+1}$, we obtain
\begin{equation}
\bfc\left(\bfK (\bfz_{k+1}^*  + h \bfK^\top \bfJ^{\top} \bflambda)\right) \approx
\bfc(\bfK \bfz_{k+1}^*)  + h \bfJ \bfK \bfK^\top \bfJ^{\top} \bflambda = \mathbf{0} ,
\label{eq:tmp2}
\end{equation}
where we neglect a second order error. This gives:
\begin{equation}
\bflambda_{k+1} \approx - h^{-1}(\bfJ \bfK \bfK^\top \bfJ^{\top})^{-1} \bfc(\bfK \bfz_{k+1}^*).
\label{eq:lambda}
\end{equation}
Inserting this expression into~\eqref{eq:update} we finally obtain
\begin{equation}
\bfz_{k+1} = \bfz_{k+1}^* -  \bfK^{\top} \bfJ^{\top} (\bfJ \bfK \bfK^\top \bfJ^{\top})^{-1} \bfc(\bfK \bfz_{k+1}^*),
\label{eq:projection_final}
\end{equation}
which performs one iteration of a linearized projection of our state onto the constraint. In practice, we perform several such projections iteratively
\begin{equation}
\bfz_{k+1}^{j+1} = \bfz_{k+1}^{j} -  \bfK^{\top} \bfJ_j^{\top} (\bfJ_j \bfK \bfK^\top \bfJ_j^{\top})^{-1} \bfc(\bfK \bfz_{k+1}^j), \quad j = 0, 1, \ldots, M,
\label{eq:projection_final_iterative}
\end{equation}
where $j$ is the projection index, 
$\bfz_{k+1}=\bfz_{k+1}^{M}$, and $\bfJ_j = \bfJ(\bfK \bfz^j_{k+1})$. The number of projections, $M$, is variable and depends upon a problem-specific early stopping criteria, as well as an upper limit. $M$ is further discussed in \Cref{sec:experiment,sec:dc}.

Based on this, the equations for a residual network similar to \eqref{resnet}, which includes projections onto the additional information given by \eqref{const} in each layer of the network can be written as
\begin{subequations}
\begin{eqnarray}
\label{presnet}
\bfz_0 &=& \bfK_o \bfx, \\
\bfz_{k+1} &=& {\rm proj}_{\bfc} \left( \bfz_k + \delta \bfq_{RK4}(\bfz_k)) \right),\ \quad k=0,1, \ldots, n-1, \\
\bfy_p  &=& \bfK \bfz_n,
\end{eqnarray}
\end{subequations}
with the projection operator defined by
\begin{eqnarray}
\label{eq:proj2}
{\rm proj}_{\bfc}(\bfz) = \bfz + \{ {\rm arg} \min_{\delta \bfz} \frac 12 \|\delta \bfz\|^2 \quad {\rm s.t.}\ \bfc(\bfK (\bfz + \delta \bfz)) = \mathbf{0} \}.
\end{eqnarray}
Here we have introduced the projection to make the network inherently obey constraints at every layer. 


Next, we note that the projection shown in~\eqref{eq:projection_final} uses Newton-like 
iterations which are accurate to second order, but can be rather expensive. 
We have found that for some problems, simple gradient descent iterations 
\begin{equation}
\label{actual_method}
\bfz_{k+1} = \bfz_{k+1}^* -  \bfK^{\top} \bfJ^{\top} \bfc(\bfK \bfz_{k+1}^*)
\end{equation}
are sufficient for our purpose. For other applications we have also applied conjugate gradient steps to approximately solve the linear system~\eqref{eq:projection_final_iterative}.

The projection methods introduced in Section~\ref{sec:end_projection} and here both allow obeying the constraints to arbitrary precision, but they can be rather expensive due to the iterative projections. 

We next introduce a faster alternative.


\subsection{Stabilization by penalty}
\label{sec:stabilization}

The last technique we explore is stabilization with respect to the additional information \eqref{const}. 
Unlike the projection, stabilization does not aim to satisfy the constraints
exactly but rather do it approximately throughout the network.
To this end, note that a descent towards the constraint can be written as an approximation
of an ODE of the form
\begin{equation*}
\dot \bfz =  -\bfK^{\top} \bfJ(\bfK \bfz)^{\top} \bfH \bfc(\bfK \bfz),
\end{equation*}
where $\bfH$ is any Symmetric Positive Definite  matrix.

Therefore, one way to stabilize the system is to augment the original ODE with a term that ``flows'' towards the constraint. 
The resulting 
limit ODE becomes
\begin{equation}
\label{stab_penalty}
\dot \bfz =  \bfg(\bfz, \bftheta) - \gamma \bfK^{\top} \bfJ(\bfK \bfz)^{\top} \bfH \bfc(\bfK\bfz).
\end{equation}
Using again an explicit method to discretize the system
we obtain a discrete analog that can be used to solve the problem.

In the following we use the RK4 discretization 
with  $\bfH=\bfI$, which 
is often sufficient. 
But for some problems more advanced choices might be needed~\cite{ascher97}. 
One common choice 
is $\bfH = (\bfJ \bfK \bfK^\top \bfJ^{\top})^{-1}$. 

{\bf Note}, in order for this approach to be numerically stable it is important to limit the maximum change a single penalty term can impose upon the data being propagated through the neural network. We limit a single penalty term to a relative change of 10\%. More details on this are provided in \Cref{app:gamma}.  


 
\section{Model Problems}
\label{sec:model_problems}

In this section we present three model problems that are used to test the different architectures.  
The multi-body pendulum described in Section~\ref{sec:multi-body-pendulum} is a simple model problem that has a well studied numerical solution~\cite{arnold2017dae}. 
The ubiquitous molecular dynamics problem described in Section~\ref{sec:water} has received much attention in the literature~\cite{allen2004introduction,Kadupitiya_2022,schutt2018schnet}. 
Finally, a vector field denoising problem with PDE constraints \cite{de2013image,fan2019brief}
is described in Section~\ref{sec:imagedenoising}.
The first two problems are viewed as constrained mechanical systems with known equations of motion that can be solved by standard integration techniques. An important difference between the two problems, though, is that the first is a simpler one where the physical constraints are precise, whereas in the second, more involved model problem the constraints are only approximate. We use numerical simulations of these physical problems to generate datasets that allow us to determine and evaluate our constrained neural networks. 
In both cases, our goal is to train a neural network to predict future states of mechanical systems given the present state. 


\subsection{The multi-body pendulum} 
\label{sec:multi-body-pendulum}

\begin{figure}[htbp]
\centering
    \def\svgscale{2}
    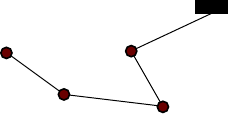
\caption{A multi-body pendulum system with four pendulums.}
\label{fig:pendulum}
\end{figure}

Our first mechanical system is a 2D multi-body pendulum, as shown in Figure~\ref{fig:pendulum}. We chose this toy experiment since it has obvious constraints, and it yields chaotic motion, which 
is non-trivial to predict. 


%
A multi-body pendulum can be parameterized in Cartesian coordinates by position 
$\bfr \in \mathbb{R}^{2\times N}$ and velocity $\bfv \in \mathbb{R}^{2\times N}$, where $N$ is the number of pendulums in the system. 
Since the distance between consecutive mass points is constant we have that at each time point
along a trajectory
\begin{equation}
    c_i = \norm{\bfr_i - \bfr_{i-1}}_2 - l_i, \quad  i=1,\ldots, N,
\label{eq:constraint_pendulum_len}
\end{equation}
where $\bfr_0=0$, and $l_i$ is the length of the $i$th pendulum piece.
The Jacobian of the constraint is easily calculated and it is a sparse matrix. In practice, only matrix-vector products with the Jacobian (and its transpose) are needed and this can be coded efficiently without storing the matrix.


\subsection{Water molecules simulation}
\label{sec:water}

For our second experiment we have created a microcanonical ensemble (NVE) simulation of 32 water molecules at a temperature of \SI{300}{K}, approximated with the Lennard-Jones force-field described in \cite{praprotnik2005molecular}, using cp2k \cite{kuhne2020cp2k}. The physical simulation is done employing a step-size of \SI{0.1}{\fs} for $100,000$ steps. The simulation is done without periodic spatial boundaries in order to have a system that fundamentally evolves over time. 

Constraints are commonly added to MD simulations in order to freeze out high frequency vibrational movement of molecules, which enables the simulations to use significantly larger time steps in physical space \cite{allen2004introduction}.

\begin{figure}[htb!]
\centering
    \def\svgscale{0.5}
    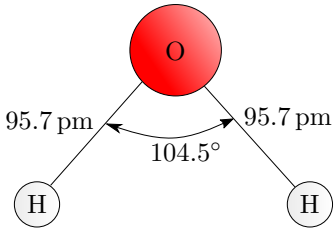
\caption{An illustration of the bound state of a water molecule.}
\label{fig:water_molecule}
\end{figure}

Water molecules are known to be bound in a triangle configuration with \SI{95.7}{\pm} between the hydrogen and oxygen atoms and an angle of \SI{104.5}{\degree} between the hydrogen atoms as illustrated in Figure~\ref{fig:water_molecule}. 
Each water molecule is constrained separately and with identical constraints
\begin{equation}
c(\bfr_i, \bfr_j) = \norm{\bfr_i - \bfr_j}_2 - l_{ij}.
\end{equation}
Note that while the multi-body pendulum data fulfills \eqref{eq:const_data}, the water molecule simulation only approximates it. For the water molecule simulation the individual molecules wiggle and vibrate which causes \eqref{eq:const_data} to only be approximately fulfilled. The fact that \eqref{eq:const_data} is not fulfilled exactly might be problematic when used in conjunction with neural networks, since it means that the solution space that the neural network is trying to emulate is not necessarily overlapping the constraint manifold that neural network output are projected onto.

\subsection{Vector field denoising}
\label{sec:imagedenoising}
Our third experiment is a vector field denoising task, where a noisy vector field is given, and the task is to denoise it. The vector field $(u,v)$ consists of randomly generated $64\times 64$ pixel images that fulfil the following divergence-free constraint:
\begin{equation}
\frac{\partial u}{\partial x} + \frac{\partial v}{\partial y} = 0.
\end{equation}
Noise is added on-the-fly, obtaining
\begin{equation*}
u_{noisy} = u + \sigma n, \ \ v_{noisy} = v + \sigma n,
\end{equation*}
where $n \sim N(0,I)$ is randomly generated using the standard normal distribution and $\sigma$ is a scalar determining the noise level. See panels (a) and (b) of~\cref{fig:imagedenoising_example} for an example.


\section{Experiments}
\label{sec:experiment}
In Section~\ref{sec:theory}, we discussed in consecutive subsections
several ways for introducing auxiliary constraint
information into a neural network, namely, \textbf{auxiliary loss},
\textbf{end constraints}, \textbf{smooth constraints}, and a \textbf{penalty} method.
Here, we next evaluate these methods on the two 
model problems discussed in the previous section.



\subsection{Multi-body pendulum}
\label{sec:experiment_pendulum}

Our first experiment is the multi-body pendulum, described in Section~\ref{sec:multi-body-pendulum}. We simulate a five-body pendulum for 100,000 steps, using the explicit RK4 discretization equation~\ref{eq:RK4}, a step size of \SI{1}{\ms}, pendulum lengths of $l_i=$\SI{1}{\m}, and pendulum masses of \SI{1}{\kg}, which ensures that the energy error over the entire simulation remains negligible. From this simulation we create our multi-body pendulum datasets.  The $i$-th data sample contains the positions and velocities of the system at steps $i$ and $i+k$. We randomly split the data into training, validation and testing datasets.
When training the neural network on the $i$-th data sample, we use the input position and velocity as initial best guess, $\bfx_i=(\bfr_i, \bfv_i)$, and try to predict the future position $\bfy_i = \bfr_{i+k}$ for a fixed value of $k$.

\begin{figure}[htb!]
	\begin{center}
    \begin{subfigure}[htb!]{0.22\textwidth}
      \includegraphics[width=\textwidth]{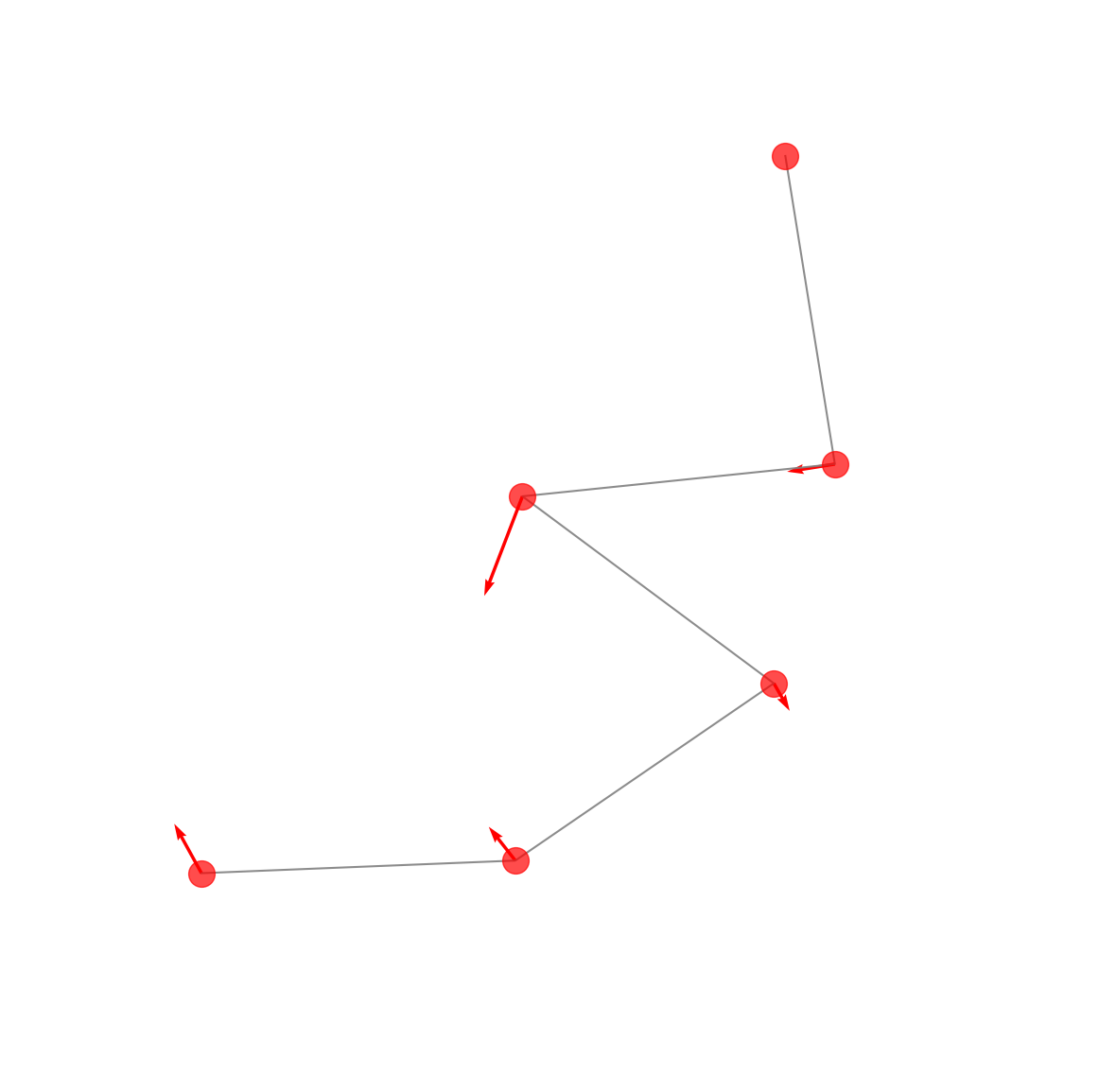}
      \caption{$t=t_0$}
      \label{fig:snapshot1}
    \end{subfigure}
    \begin{subfigure}[htb!]{0.22\textwidth}
      \includegraphics[width=\textwidth]{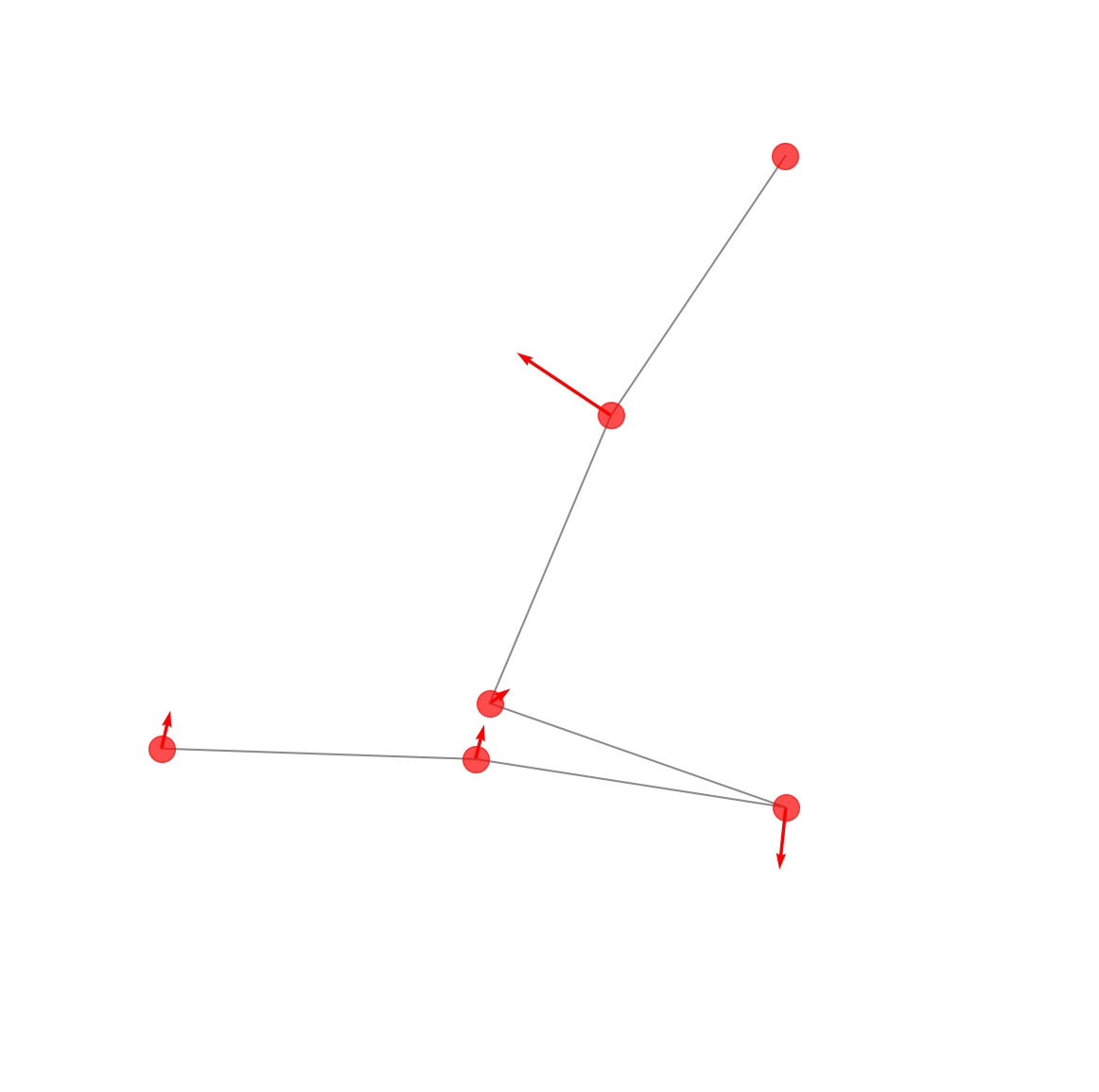}
      \caption{$t=t_0 + \SI{0.1}{\s}$}
      \label{fig:snapshot2}
    \end{subfigure}
    \begin{subfigure}[htb!]{0.22\textwidth}
      \includegraphics[width=\textwidth]{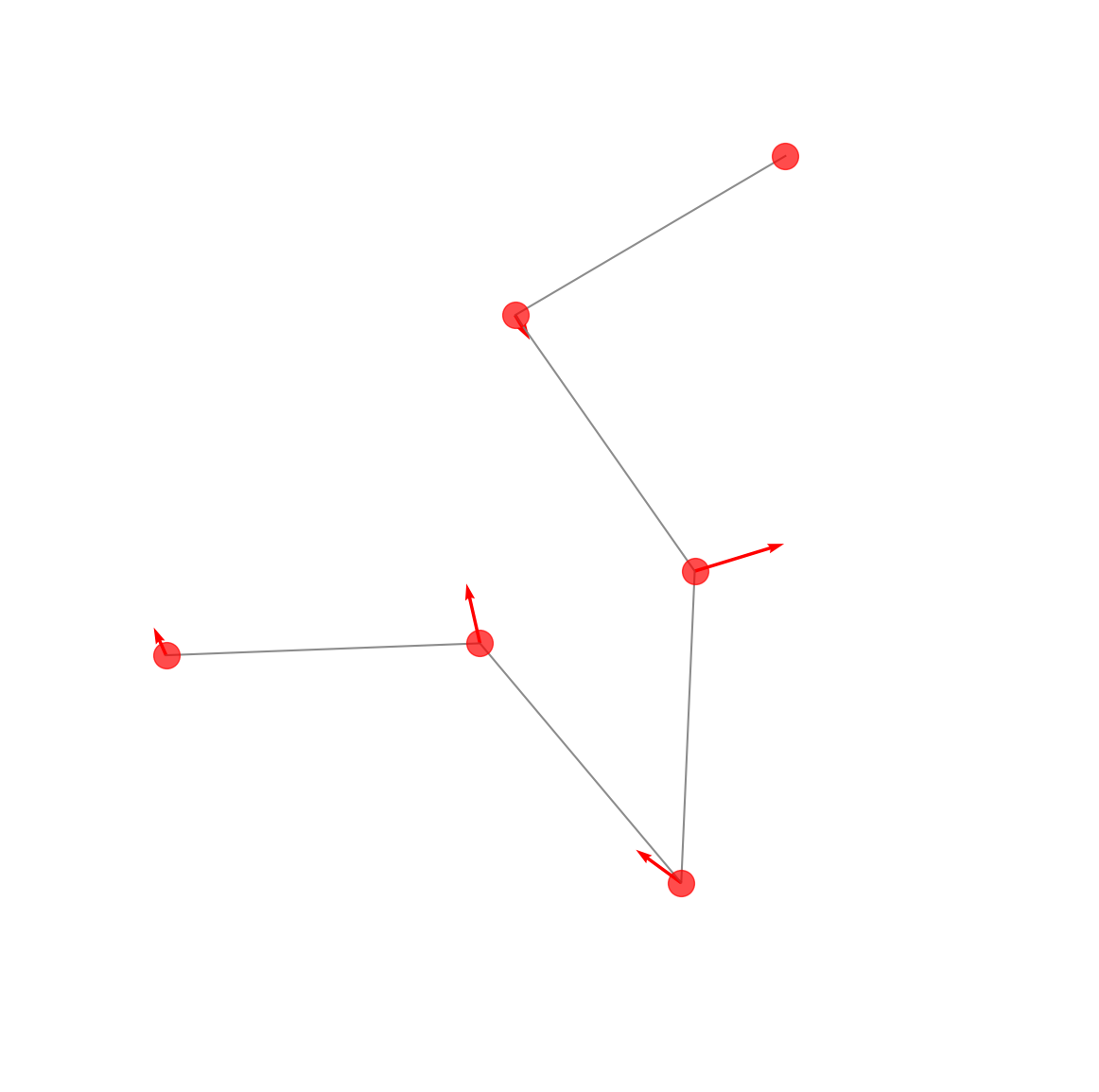}
      \caption{$t=t_0 + \SI{0.2}{\s}$}
      \label{fig:snapshot3}
    \end{subfigure}
    \begin{subfigure}[htb!]{0.22\textwidth}
      \includegraphics[width=\textwidth]{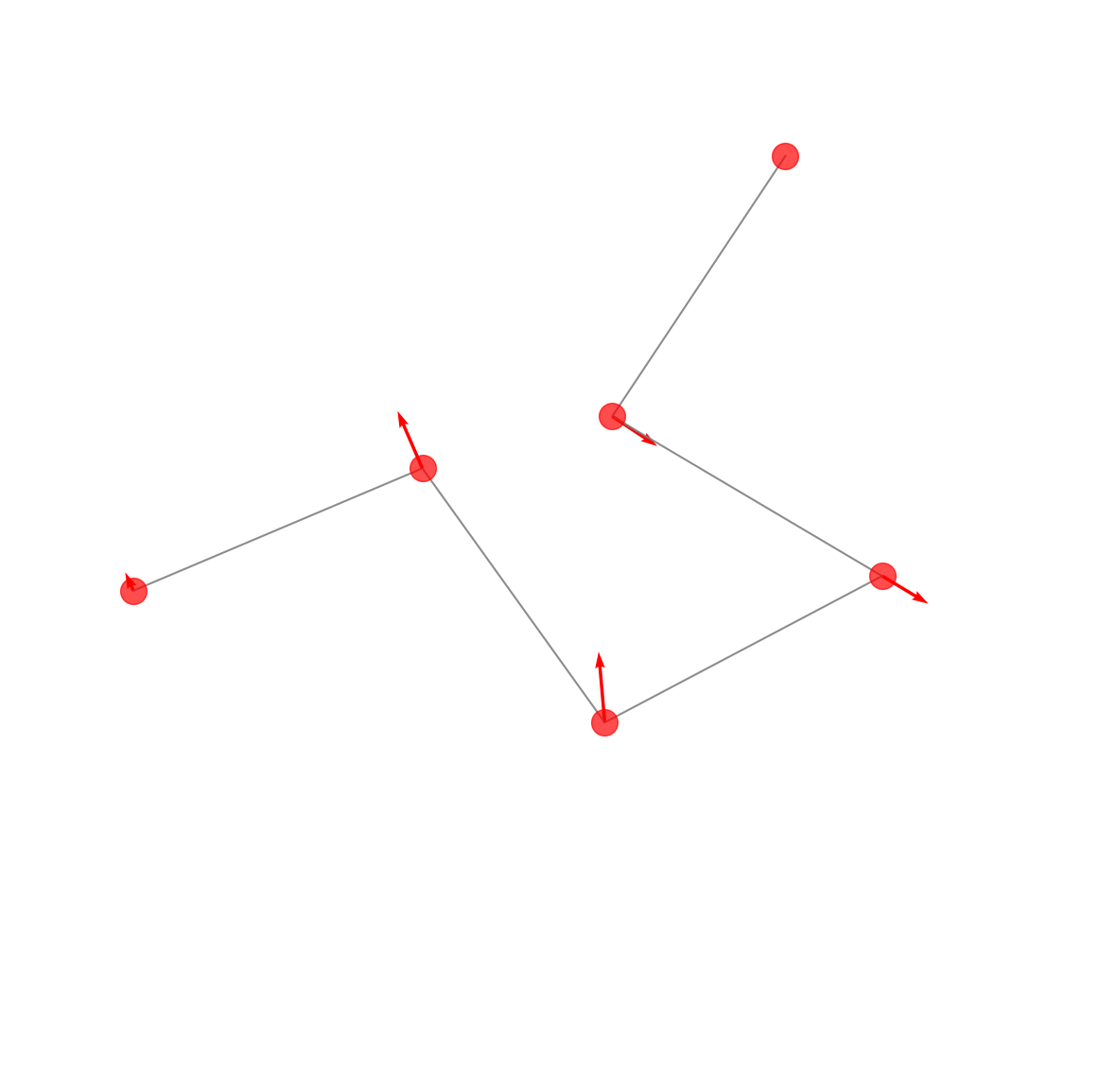}
      \caption{$t=t_0 + \SI{0.3}{\s}$}
      \label{fig:snapshot4}
    \end{subfigure}
    \end{center}
\caption{Snapshots of a five-body pendulum. These should be viewed as snapshots of a single pendulum animation. Each snapshot is 100 steps after the previous one (\SI{0.1}{\s}). The arrows indicate the velocity of each individual pendulum.}
\label{fig:pendulum_snapshots}
\end{figure}

We implement our neural networks using RK4 
stepping between the layers and a total of 8 layers. The neural  network is trained with a batch-size of 10. For the constraints, we use at most 200 gradient descent projections to obey the constraints with an early stopping criterion if the max constraint violation is ever below \SI{e-4}{\m}. The hyperparameter values were found using a simple grid search.
In this experiment we can control how difficult the problem is, by varying $k$.
Figure~\ref{fig:pendulum_snapshots} shows 4 snapshots from our five-pendulum simulation configuration (the ground truth).

\begin{figure}[htb!]
	\begin{center}
      \includegraphics[width=0.45\textwidth]{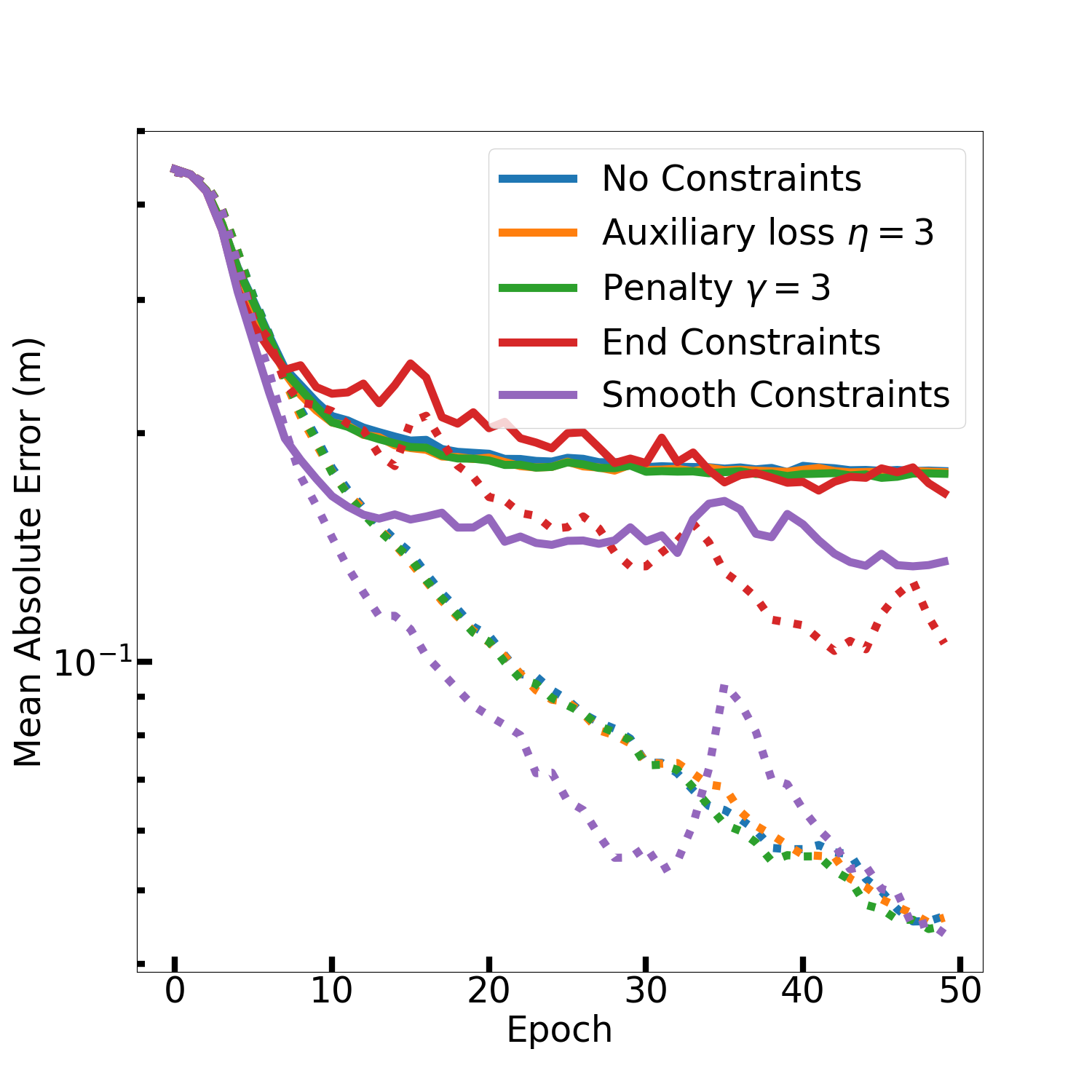}
      \includegraphics[width=0.45\textwidth]{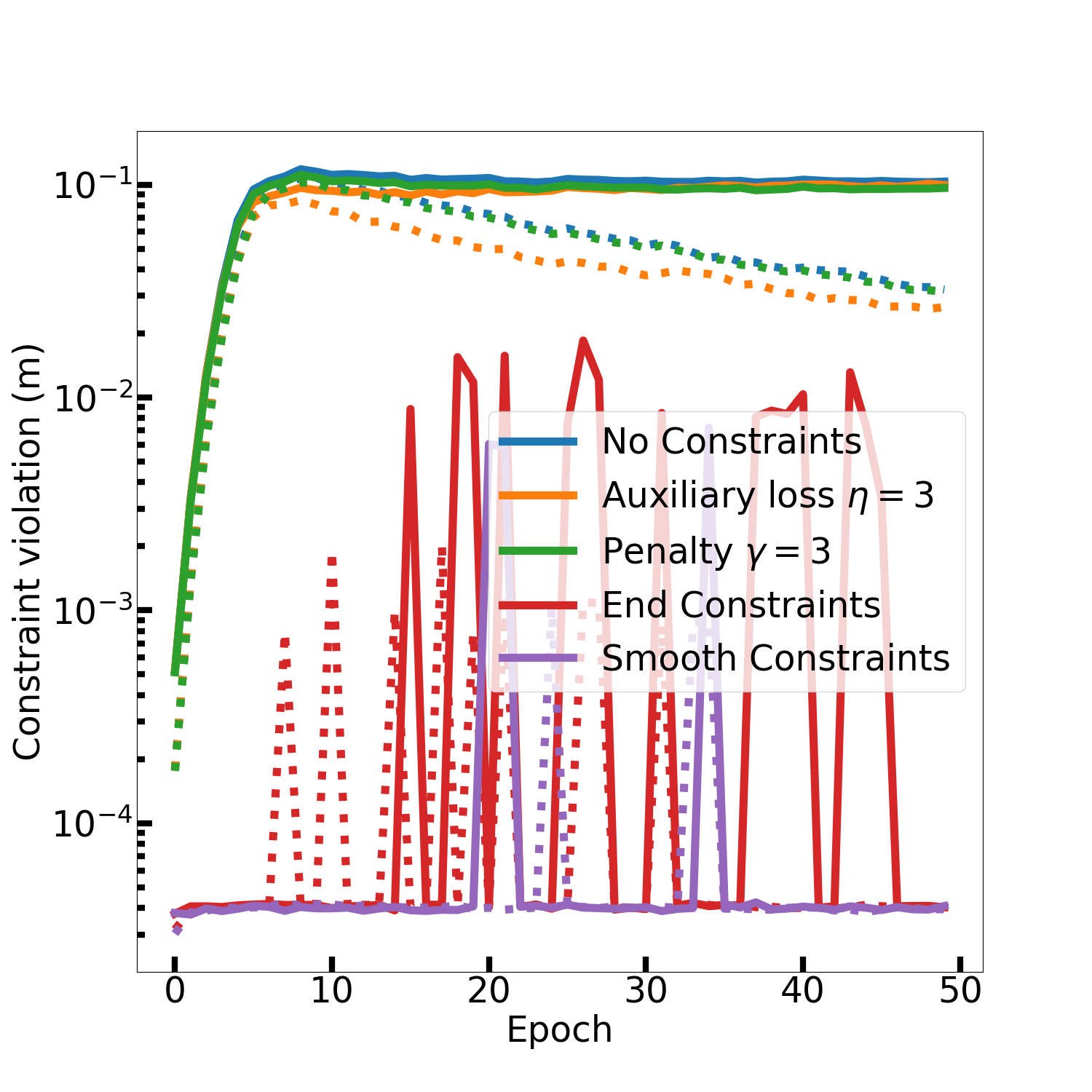}
    \end{center}
\caption{Comparison of the different ways of adding constraints to a neural network trained on 100 multi-body pendulum samples with $k=100$. Left: Mean absolute error. Right: Maximum constraint violation. Each run has been repeated three times. The solid/dashed lines are the average based on validation/training data. Note the large fluctuations in performance for the projection methods (End constraints and Smooth Constraints), which shows clear signs of instabilities.}
\label{fig:npendulum_initial_test}
\end{figure}

We start by determining appropriate values for $\gamma$ and $\eta$ used in the penalty method 
and the auxiliary regularization method 
(For more details see Appendices~\ref{app:gamma}-\ref{app:regularization}).
We then compare the different ways of adding constraints to a neural network. 
Figure~\ref{fig:npendulum_initial_test} shows the results of training a mimetic neural network without and with constraints \cite{eliasof2021mimetic}. 

Based on these results, we can see that end constraints and smooth constraints methods lead to unstable training, which is due to the fact that the neural network no longer predicts solutions even remotely close to obeying the constraints. Instead the solutions are projected onto the constraints, which works well for small constraint violations, but leads to instabilities for large constraint violations. In order to counteract this, we want the network to still predict reasonable pendulum positions before constraints are applied, and to that effect we can apply two different methods. One is to also add a penalty term to the end- and smooth-constraints, which naturally leads the network towards a solution that is close to obeying the constraints, as described in Section~\ref{sec:stabilization}. The second method is to add an auxiliary regularization term to the loss function as described in Section~\ref{sec:regularization} based on the constraint violation before any projections onto the constraints are applied in the neural network. 

\begin{figure}[htb!]
	\begin{center}
      \includegraphics[width=0.45\textwidth]{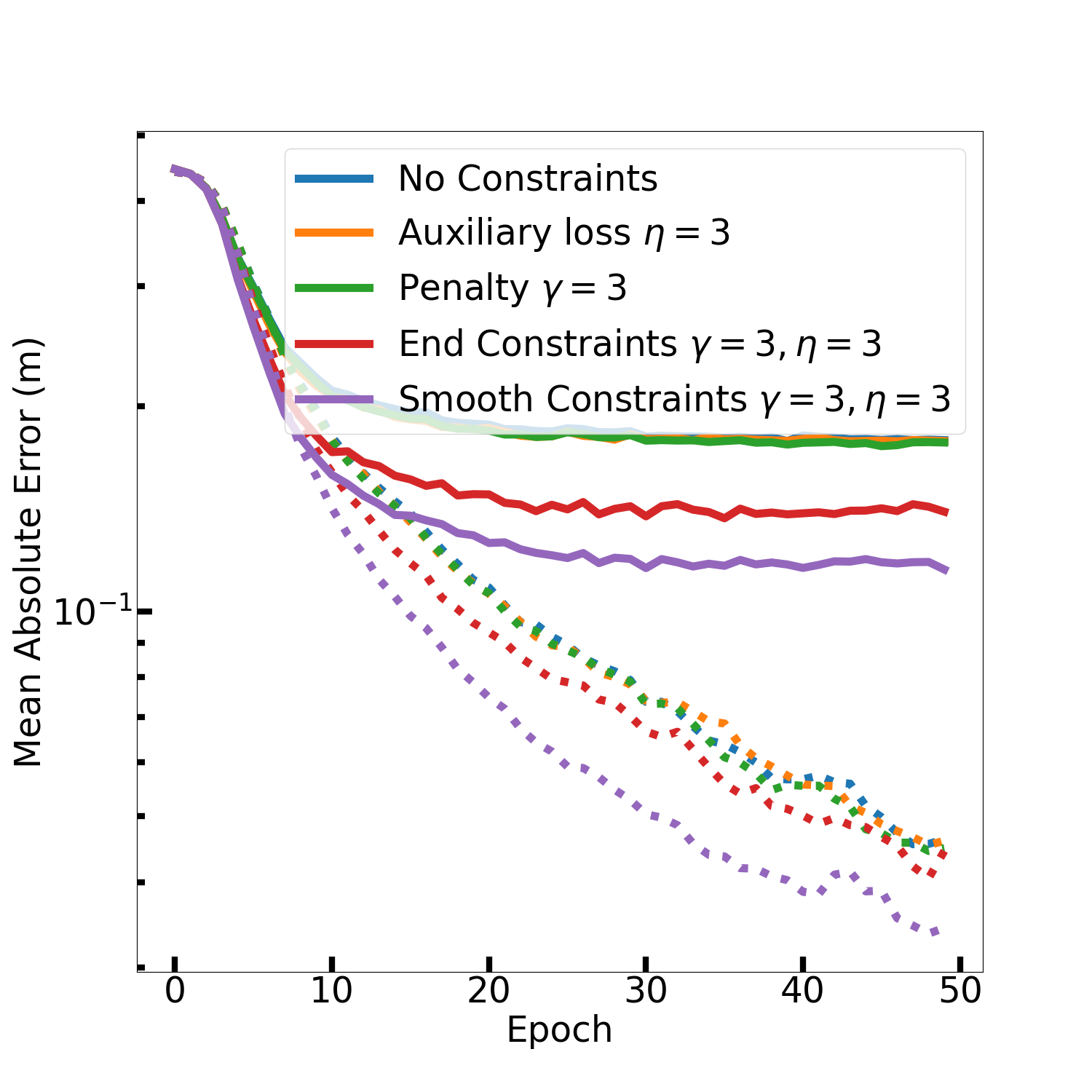}
      \includegraphics[width=0.45\textwidth]{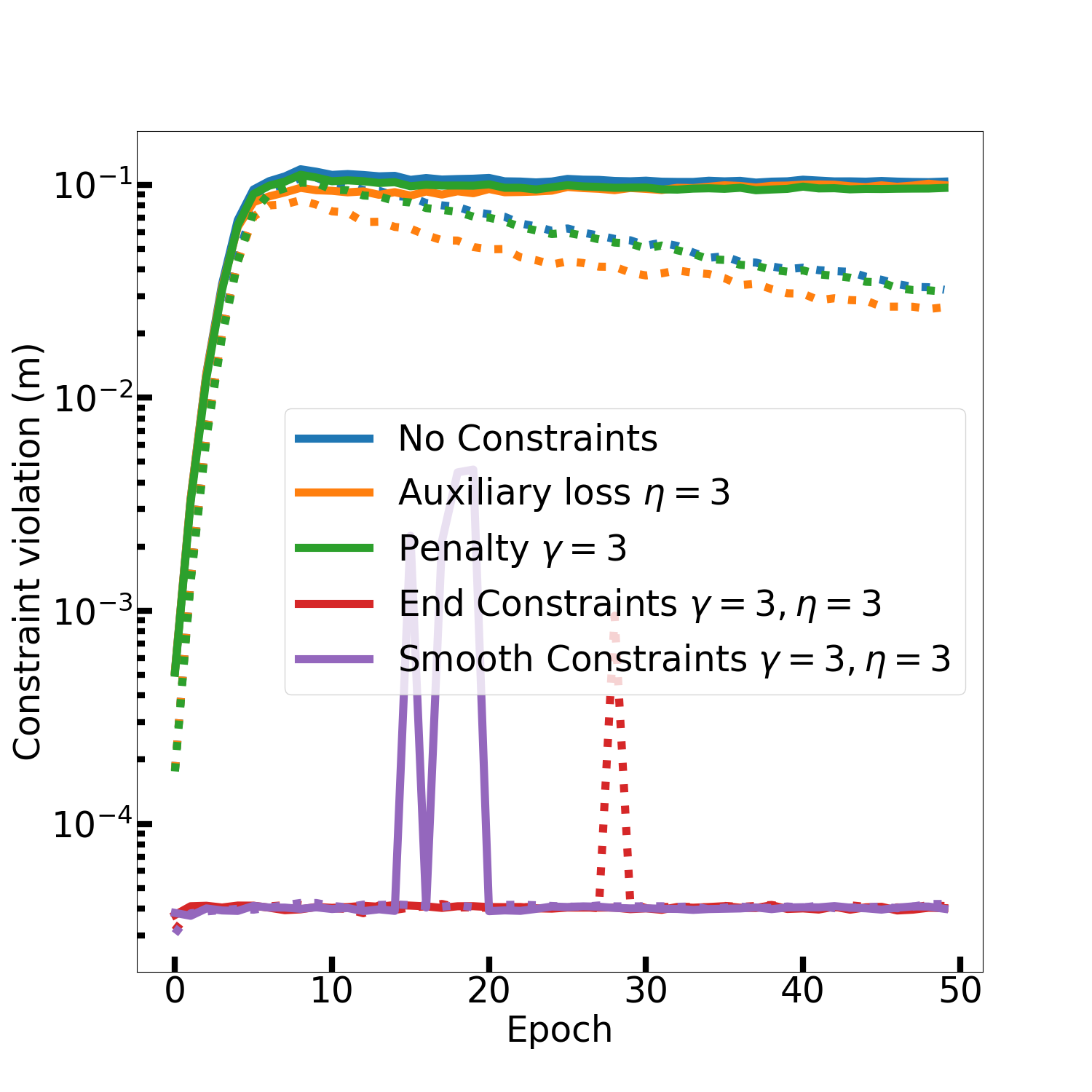}
    \end{center}
\caption{Comparison of the different ways of adding constraints to neural network trained on 100 multi-body pendulum samples with $k=100$. Left: Mean absolute error. Right: Maximum constraint violation. Each run has been repeated three times, the solid/dashed lines are the average based on validation/training data. The blue and orange can be hard to see, but is hidden underneath the green. Note that the various methods are able to train in roughly the same number of epochs, but the stabilized projection methods reach a significantly lower error, especially on the validation data.}
\label{fig:npendulum_final_test}
\end{figure}

We test both alternatives separately with various strengths, and also a combination of the two, and find that the penalty alternative is insufficient to completely stabilize the training. Auxiliary loss on the other hand does stabilize the network if applied with sufficient strength. For both end constraints, and smooth constraints, we find that a combination of auxiliary loss and penalty gives the best result. We still refer to these combined methods as end constraints or smooth constraints, but specify a value for $\gamma,\eta$. 
With the modified smooth and end constraints, we rerun the initial comparative experiment as shown in Figure~\ref{fig:npendulum_final_test}.

\begin{figure}[htb!]
	\begin{center}
    \begin{subfigure}[htb!]{0.18\textwidth}
      \includegraphics[width=\textwidth]{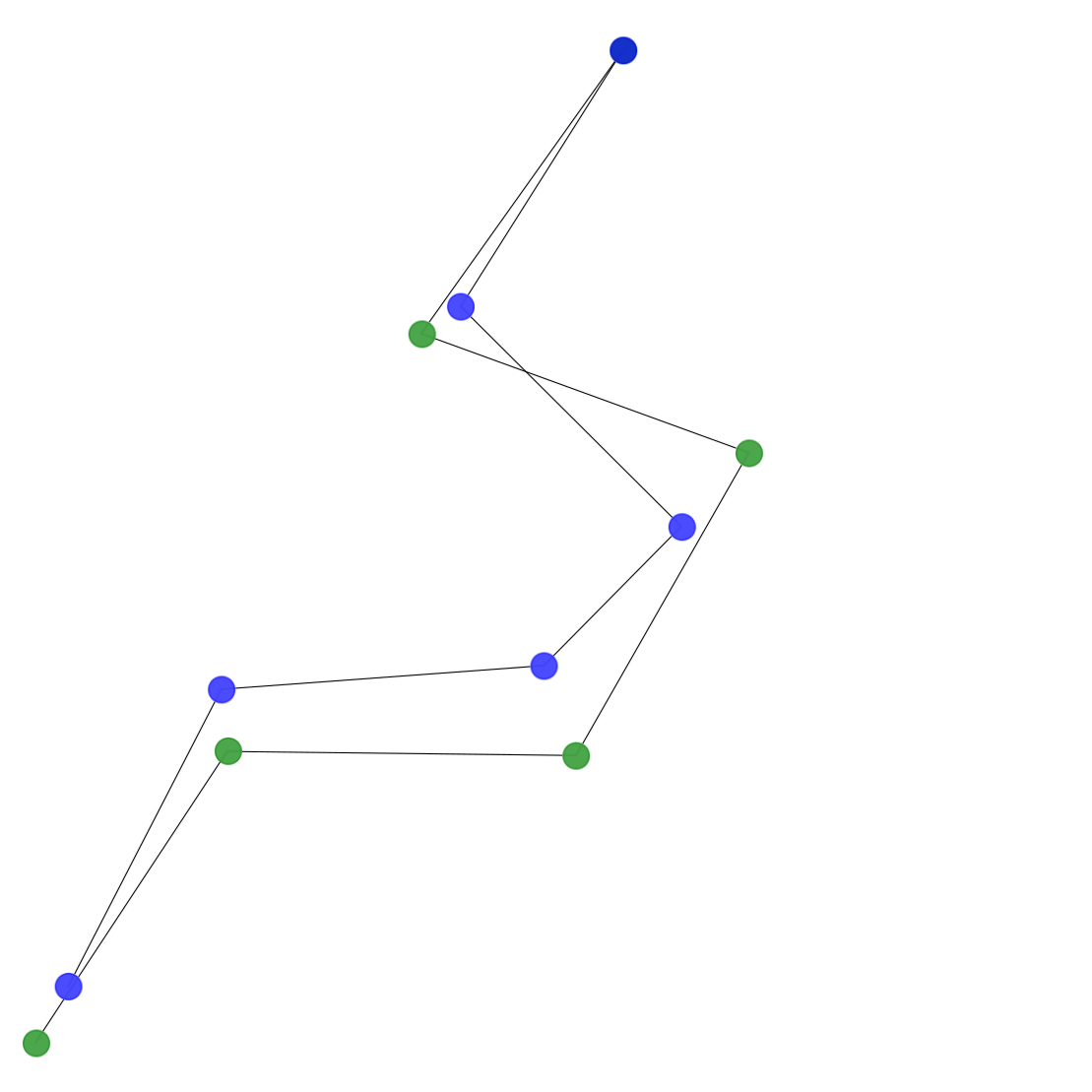}
      \caption{No con\\ \centerline{21.2}}
    \end{subfigure}
	\begin{subfigure}[htb!]{0.18\textwidth}
      \includegraphics[width=\textwidth]{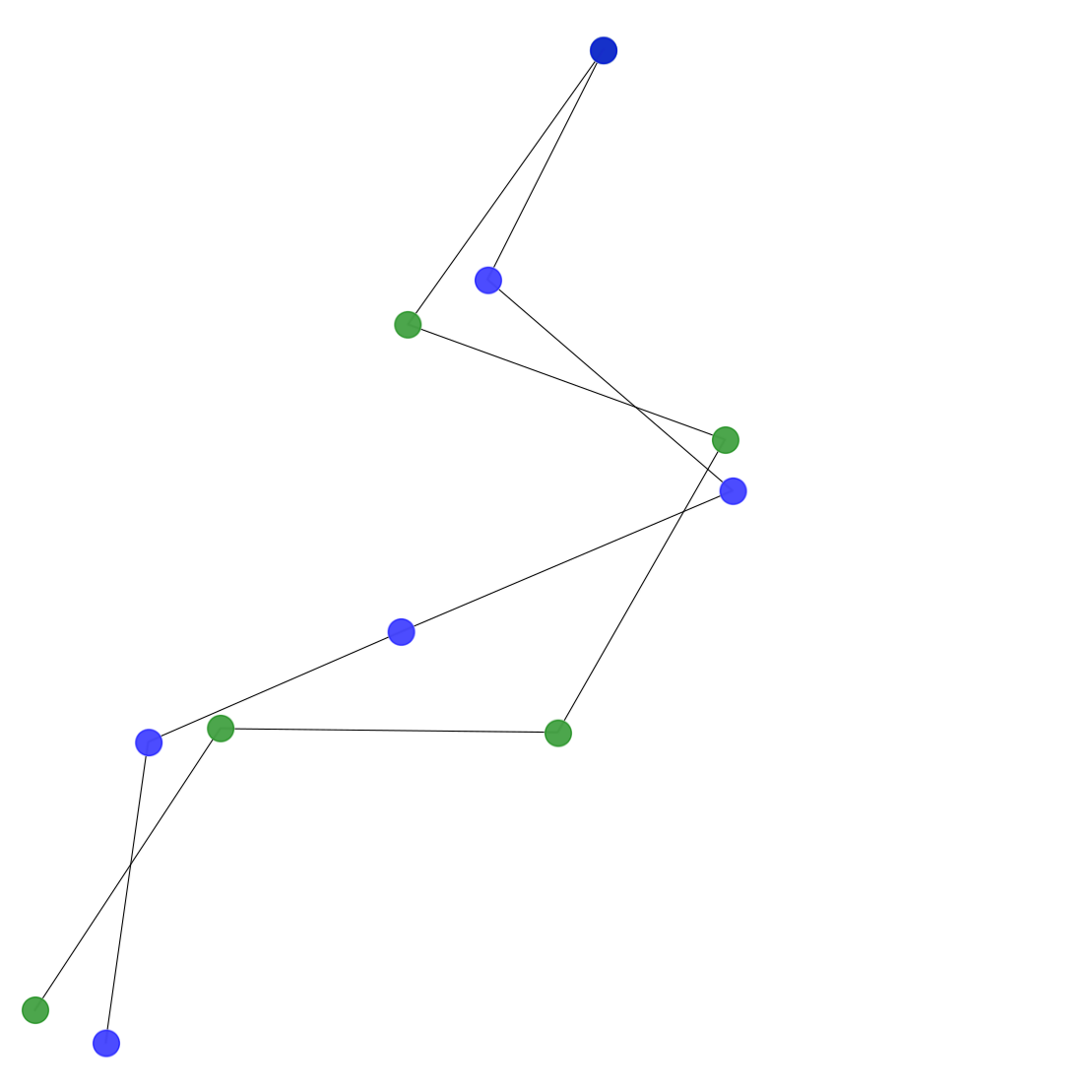}
      \caption{Aux con\\ \centerline{28.5}}
    \end{subfigure}
    \begin{subfigure}[htb!]{0.18\textwidth}
      \includegraphics[width=\textwidth]{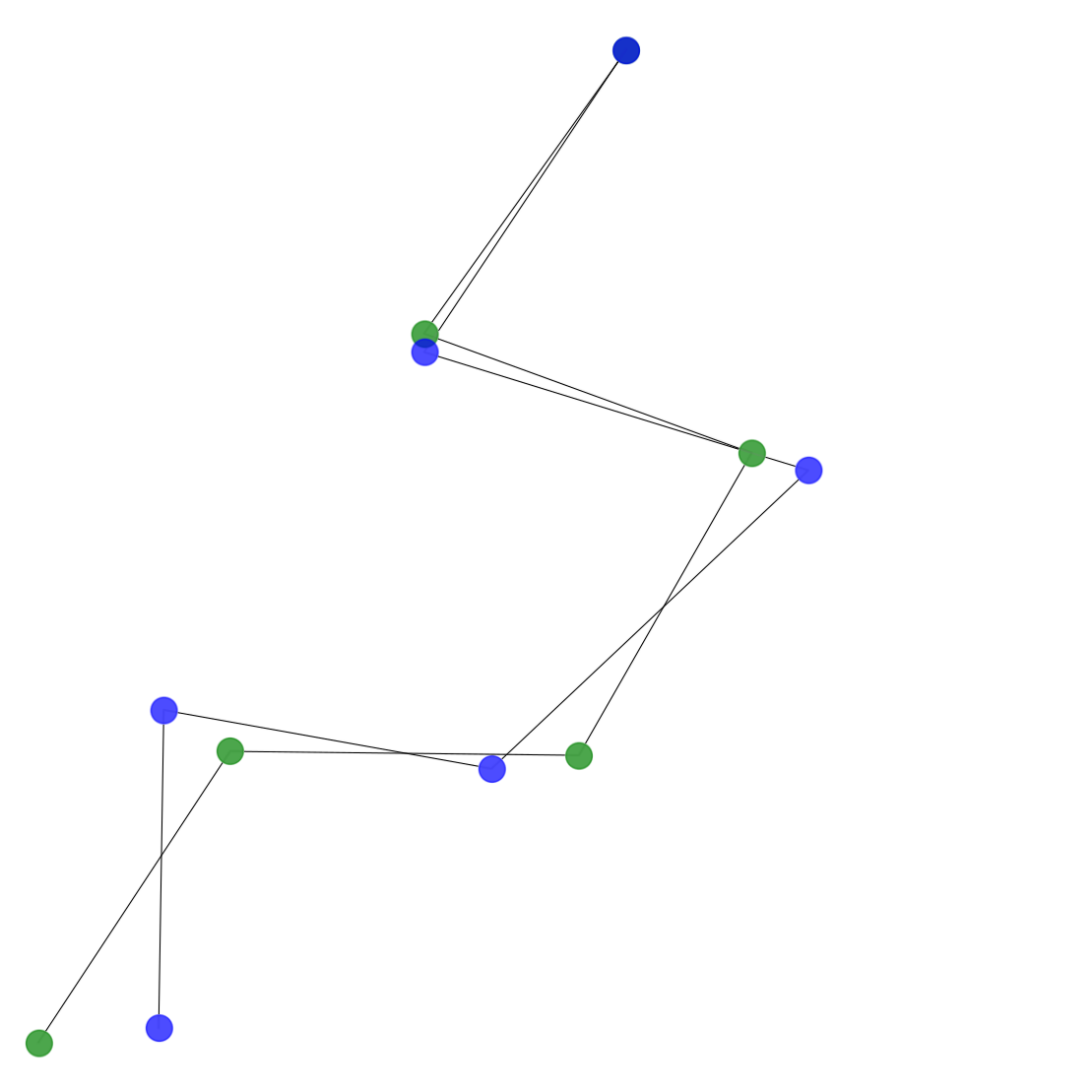}
      \caption{Penalty\\ \centerline{21.0}}
    \end{subfigure}
    \begin{subfigure}[htb!]{0.18\textwidth}
      \includegraphics[width=\textwidth]{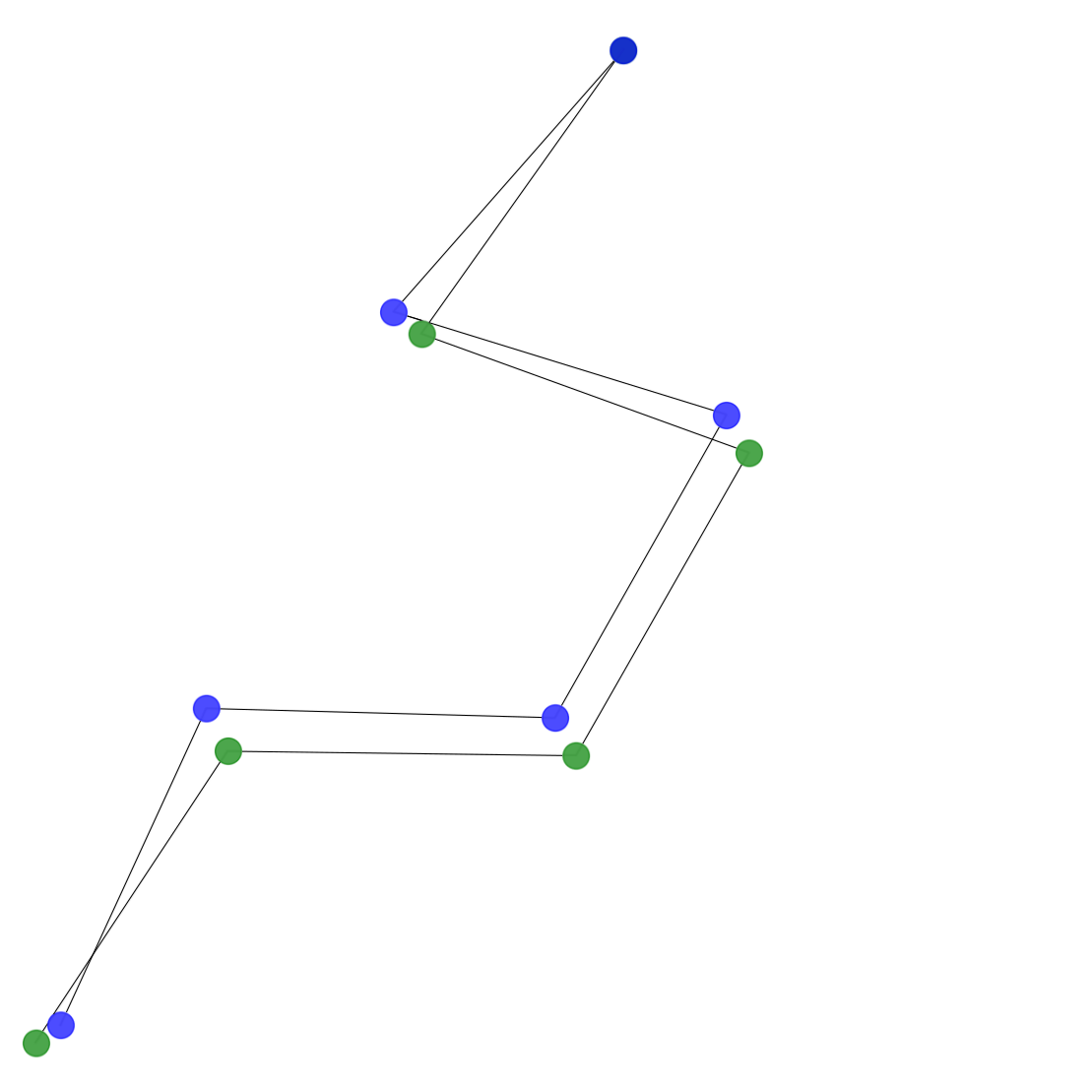}
      \caption{End con\\ \centerline{11.5}}
    \end{subfigure}
    \begin{subfigure}[htb!]{0.18\textwidth}
      \includegraphics[width=\textwidth]{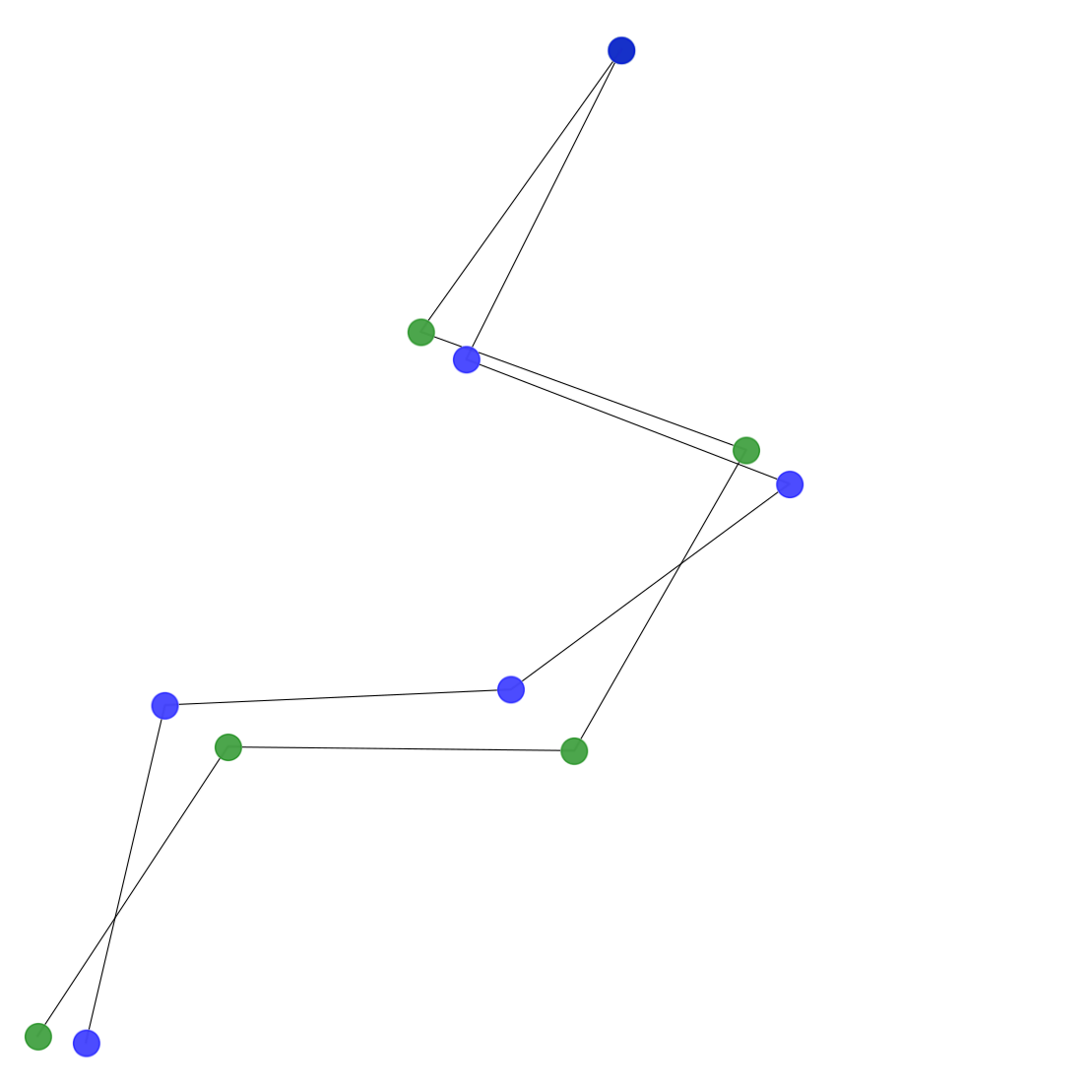}
      \caption{Smooth con\\ \centerline{18.5}}
    \end{subfigure}
    \end{center}
\caption{Comparative snapshots of the different trained networks during test evaluation. The networks are trained with $k=200$ on 1000 training samples. The green pendulum is the desired output, while the blue pendulums are the predicted output. The number given below each pendulum is the mean absolute error (\SI{}{\cm}) for that particular prediction. Note how the unconstrained pendulum gives a reasonable result, but despite this it still clearly violates the pendulum constraint (the length of pendulum three is clearly significantly shorter than \SI{1}{\m}).}
\label{fig:npendulum_final_snapshots}
\end{figure}

We have investigated the effect of training sample size as well as the difficulty of the problem and show the results of this in Table~\ref{tab:pendulum}. Figure~\ref{fig:npendulum_final_snapshots} shows a comparative example of a multi-body pendulum prediction based of neural networks with different constraining schemes. Additional training/validation comparisons can be seen in Appendix~\ref{app:additional_examples}

\begin{table}
\begin{center}
\begin{tabular}{ l|l|r|r|r|r|r|r }
& Constraints & \multicolumn{2}{c|}{$n_t=100$} & \multicolumn{2}{c|}{$n_t=1000$} & \multicolumn{2}{c}{$n_t=10000$} \\
&  & MAE & CV & MAE & CV & MAE & CV \\
\hline
\multirow{5}{*}{$k=100$}
& No constraints & 17.7 & 10.3 & 7.39 & 5.06 & 1.54 & 1.15 \\
& Auxiliary loss $\eta=3$ & 17.7 & 7.41 & 7.77 & 4.35 & 1.51 & 0.94 \\
& Penalty $\gamma=3$ (ours) & 17.5 & 9.65 & 7.01 & 3.99 & 1.63 & 0.96   \\
& End constraints $\gamma,\eta=3$ (ours) & 14.0 & \textbf{0.00} & 5.70 & \textbf{0.00} & 1.29 & \textbf{0.00} \\
& Smooth constraints $\gamma,\eta=3$ (ours) & \textbf{11.2} & \textbf{0.00} & \textbf{3.35} & \textbf{0.00} & \textbf{1.02} & \textbf{0.00} \\
\hline
\multirow{5}{*}{$k=200$}
& No constraints & 46.1 & 25.2 & 26.1 & 13.4 & 3.26 & 2.30\\
& Auxiliary loss $\eta=1$ & 46.1 & 20.5 & 25.9 & 10.4 & 3.32 & 2.03 \\
& Penalty $\gamma=1$ (ours) & 45.5 & 24.2 & 25.9 & 13.1 & 3.42 & 2.34 \\
& End constraints $\gamma,\eta=1$ (ours) & 42.6 & 6.91 & 27.8 & 2.66 & 3.04 & \textbf{0.00} \\
& Smooth constraints $\gamma,\eta=1$ (ours) & \textbf{33.3} & \textbf{0.00} & \textbf{19.1} & \textbf{0.00} & \textbf{2.21} & \textbf{0.00} \\
\end{tabular}
\end{center}
\caption{Mean absolute error (MAE) and constraint violation (CV), in \SI{}{\cm}, over a test set of 1000 samples on the multi-body pendulum problem. $n_t$ is the number of training samples, while $k$ is the number of steps predicted ahead. Each experiment was repeated three times and the average of those runs are shown. In all experiments we see that the projection methods generally leads to better results than any of the alternatives with smooth constraints being the superior method by a significant margin.}
\label{tab:pendulum}
\end{table}


\subsection{Molecular dynamics simulations - water molecules}

Our second experiment is the water molecule simulation described in Section~\ref{sec:water}. Similarly to the multi-body pendulum experiment, the $i$-th data sample contains positions and velocities of the system at $i$ and $i+k$ steps.
At the start of each experiment we randomly generate training/validation/testing datasets based on the $100,000$ physical time steps already simulated. When training the neural network on the $i$th data sample, we use the input position and velocity vectors as initial best guess, $\bfx_i = (\bfr_i,\bfv_i)$, and try to predict the future position $\bfy_i = \bfr_{i+k}$.  
Due to the nature of the simulation we change from a mimetic neural network to an SO3-equivariant neural network using the e3nn framework \cite{batzner20223}. The neural network is trained with a batch-size of 10. For the projection constraints, we use a maximum of 100 steepest descent projections and an early stopping of \SI{5e-2}{\pm}. The hyperparameter values were found using a simple grid search.  

\begin{figure}[htb!]
	\begin{center}
      \includegraphics[width=0.45\textwidth]{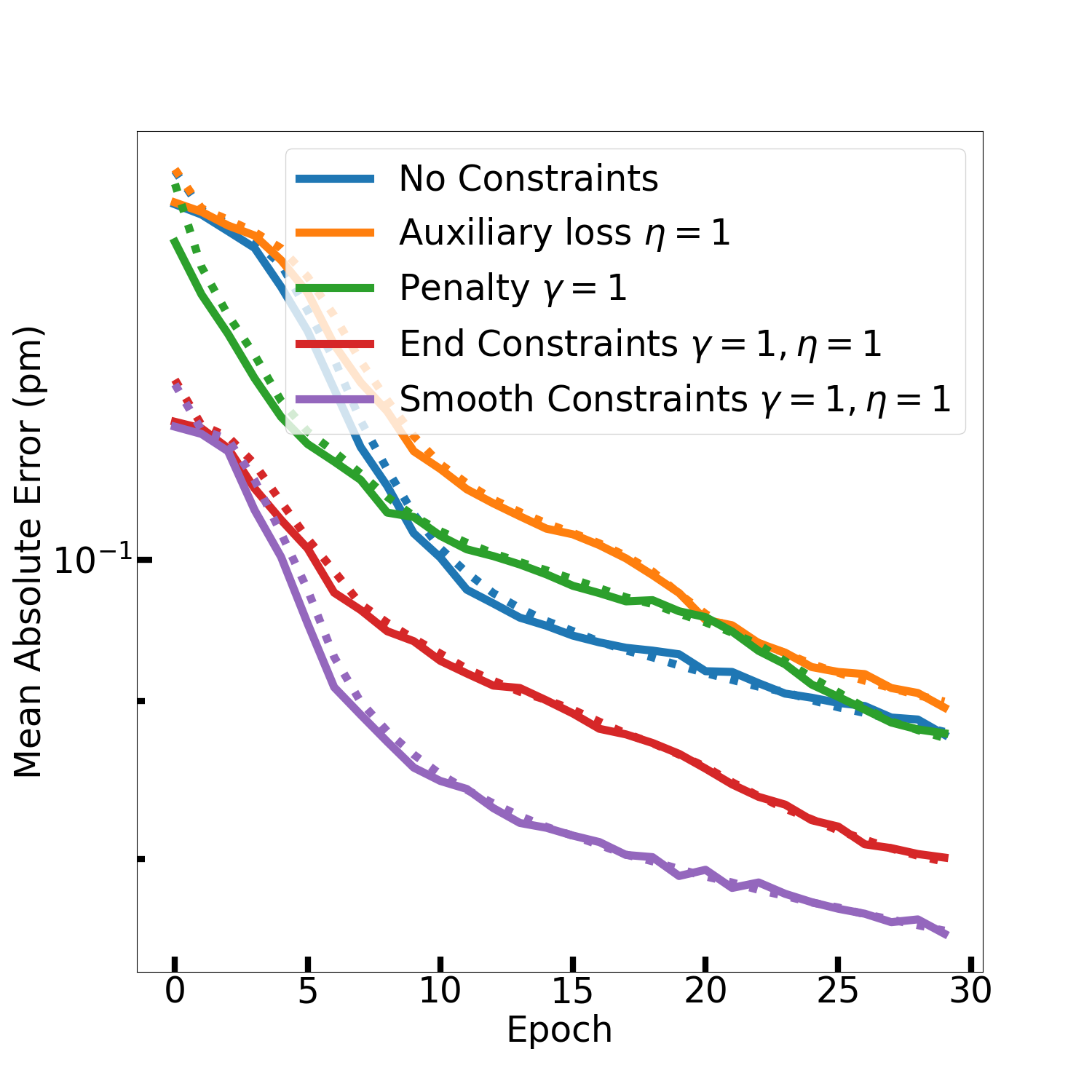}
      \includegraphics[width=0.45\textwidth]{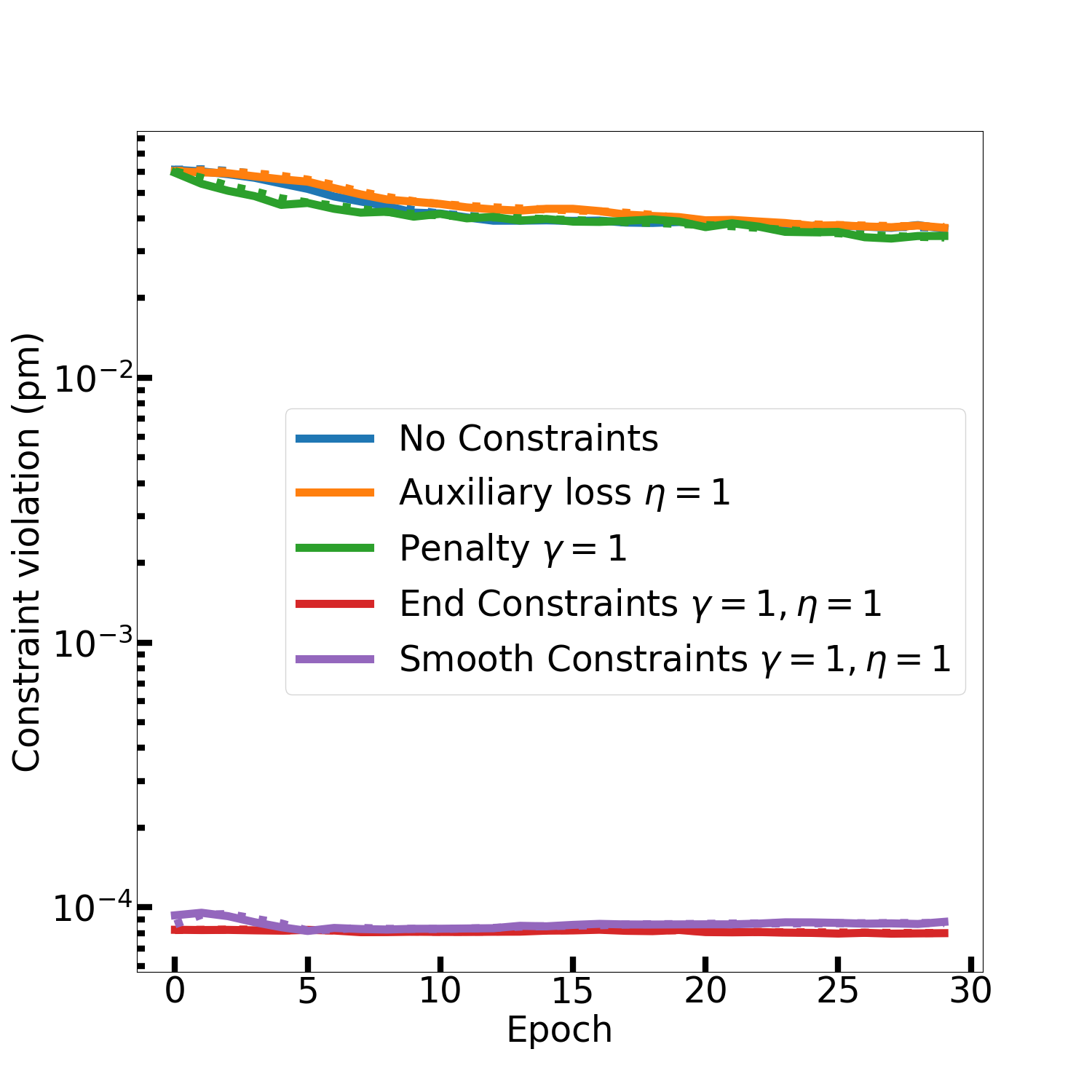}
    \end{center}
\caption{Comparison of the different ways of adding constraints to neural network trained on 10000 water molecule samples with $k=50$. Left: Mean absolute error. Right: Maximum constraint violation. Each run has been repeated three times, the solid/dashed lines are the average based on validation/training data. The blue line can be hard to see, since it is mostly hidden beneath the orange line. Note that for this problem the projection methods already from the first epoch reaches a significantly lower error.}
\label{fig:water_10000}
\end{figure}

Once again, we use the modified versions of smooth-constraints and end-constraints. Training and validation with 10000 samples can be seen in Figure~\ref{fig:water_10000}, while Table~\ref{tab:water} shows a comparison of the different constraint methods with various sizes of training data. 

\begin{table}
\begin{center}
\begin{tabular}{ l|r|r|r|r|r|r }
Constraints & \multicolumn{2}{c|}{$n_t=100$} & \multicolumn{2}{c|}{$n_t=1000$} & \multicolumn{2}{c}{$n_t=10000$} \\
& MAE & CV & MAE & CV & MAE & CV \\
\hline
No constraints & 13.1 & 6.21 & 12.8 & 5.99 & 8.78 & 3.71 \\
Auxiliary loss $\eta=1$ & 13.2 & 6.14 & 12.9 & 5.93 & 8.97 & 3.69 \\
Penalty $\gamma=1$ (ours) & 13.1 & 6.19 & 11.9 & 5.16 & 8.78 & 3.37 \\
End constraints $\gamma,\eta=1$ (ours) & 11.2 & \textbf{0.01} & \textbf{11.0} & \textbf{0.01} & 8.02 & \textbf{0.01} \\
Smooth constraints $\gamma,\eta=1$ (ours) & \textbf{11.1} & \textbf{0.01} & \textbf{11.0} & \textbf{0.01} & \textbf{7.58} & \textbf{0.01} \\
\end{tabular}
\end{center}
\caption{Mean absolute error (MAE) and constraint violation (CV), in \SI{}{\pm}, over a test set of 1000 samples on the water molecule problem. $n_t$ is the number of training samples. $k=50$ for all experiments. 
Each experiment was repeated three times and the average of those runs are shown. In all experiments we see that the projection methods (End constraints and Smooth constraints) gives significantly better results than any of the alternatives. }
\label{tab:water}
\end{table}


\subsection{Vector field denoising}
Our third experiment is the vector field denoising described in Section~\ref{sec:imagedenoising}. 300 vector fields were generated before the experiments start, 100 fields are assigned to training, 100 to validation, and 100 to testing. As a neural network, we used a standard convolutional residual neural network with 4 layers and 205,152 parameters. Training was performed with a learning rate of $10^{-3}$ and a batch size of 10. For the projection constraints, we use a maximum of 100 steepest descent projections and an early stopping of $10^{-3}$. The hyperparameter values were found using a simple grid search. All models were trained using fields with noise level $\sigma=1$.  

\begin{figure}
\setlength\tabcolsep{1.5pt}
\begin{tabular}{ccccccc}
  \includegraphics[height=17mm]{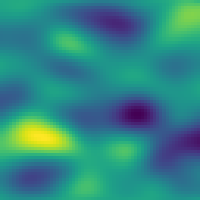} &
  \includegraphics[height=17mm]{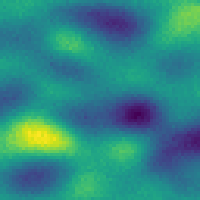} \setlength\tabcolsep{5pt} &
  \includegraphics[height=17mm]{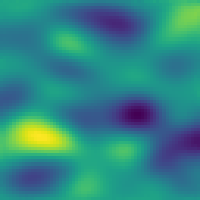} &
  \includegraphics[height=17mm]{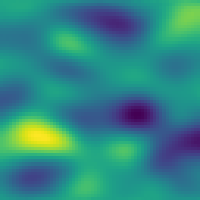} &
  \includegraphics[height=17mm]{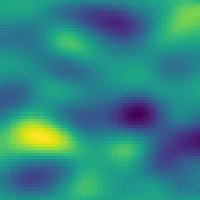} &
  \includegraphics[height=17mm]{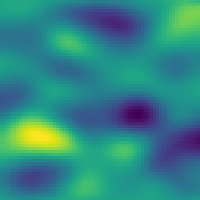} &
  \includegraphics[height=17mm]{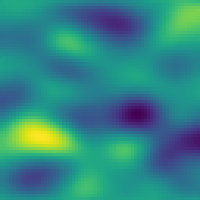}   \\
  \includegraphics[height=17mm]{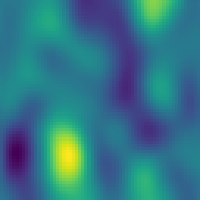} &
  \includegraphics[height=17mm]{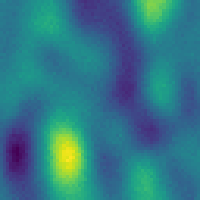} \setlength\tabcolsep{5pt} &
  \includegraphics[height=17mm]{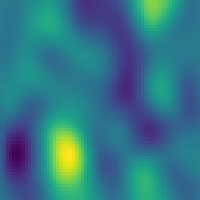} &
  \includegraphics[height=17mm]{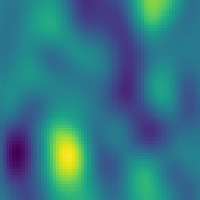} &
  \includegraphics[height=17mm]{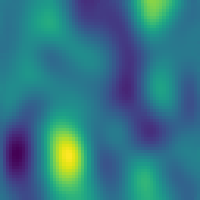} &
  \includegraphics[height=17mm]{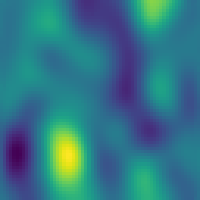} &
  \includegraphics[height=17mm]{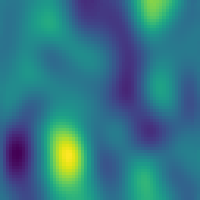} \\
  (a) & (b) & (c) & (d) & (e) & (f) & (g) 
\end{tabular}
\caption{Vector field denoising example: (a) true vector field, top row for $u$, bottom row for $v$, (b) the noisy input to the neural networks; 
(c)-(g) are the denoised vector fields from neural networks: (c) no constraints, (d) regularization $\gamma=1$, (e) penalty $\eta=1$, (f) end constraints $\gamma=0.1$, (g) smooth constraints $\gamma=0.01$. Training was done with $\sigma=1$.}
\label{fig:imagedenoising_example}
\end{figure}

\begin{figure}[htb!]
	\begin{center}
      \includegraphics[width=0.45\textwidth]{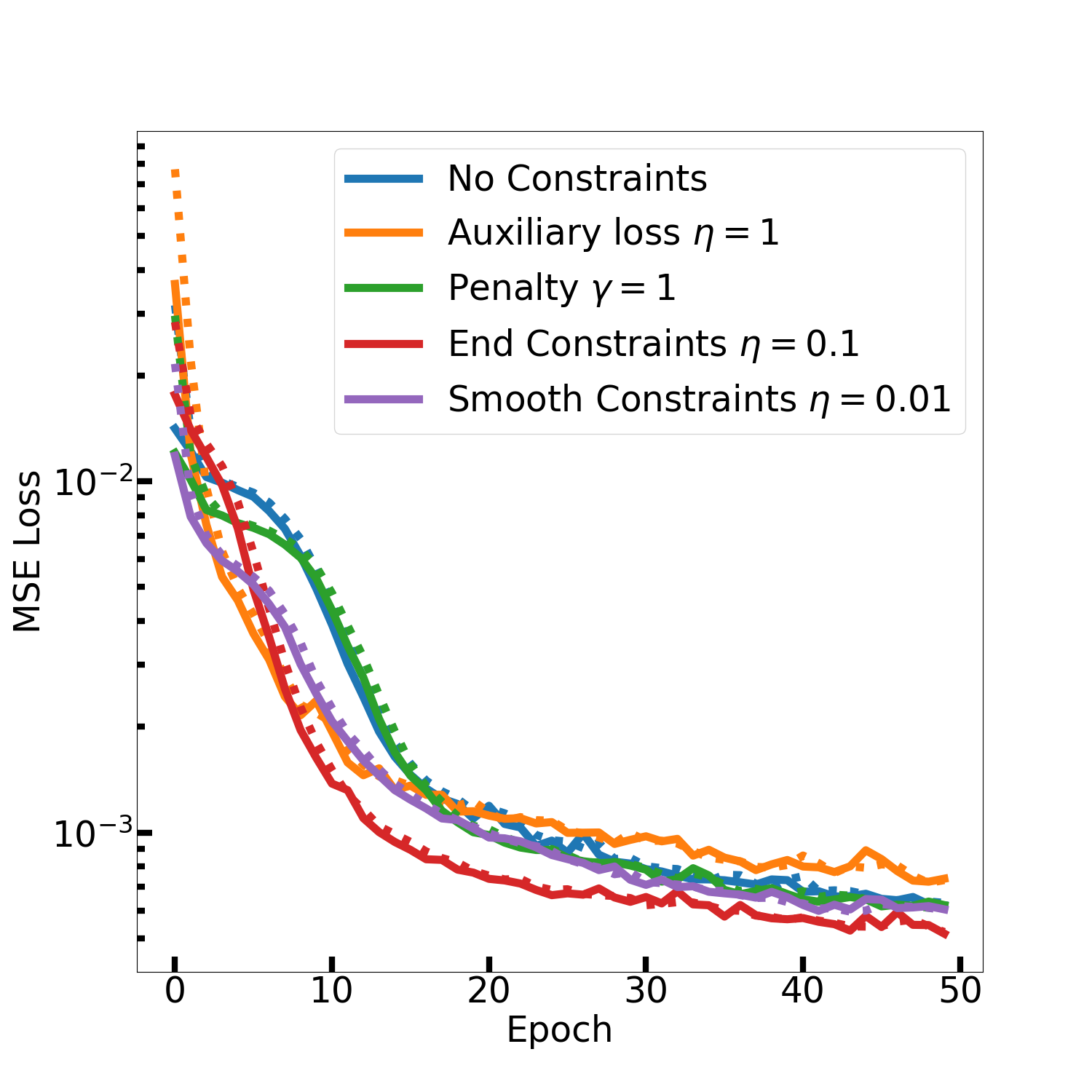}
      \includegraphics[width=0.45\textwidth]{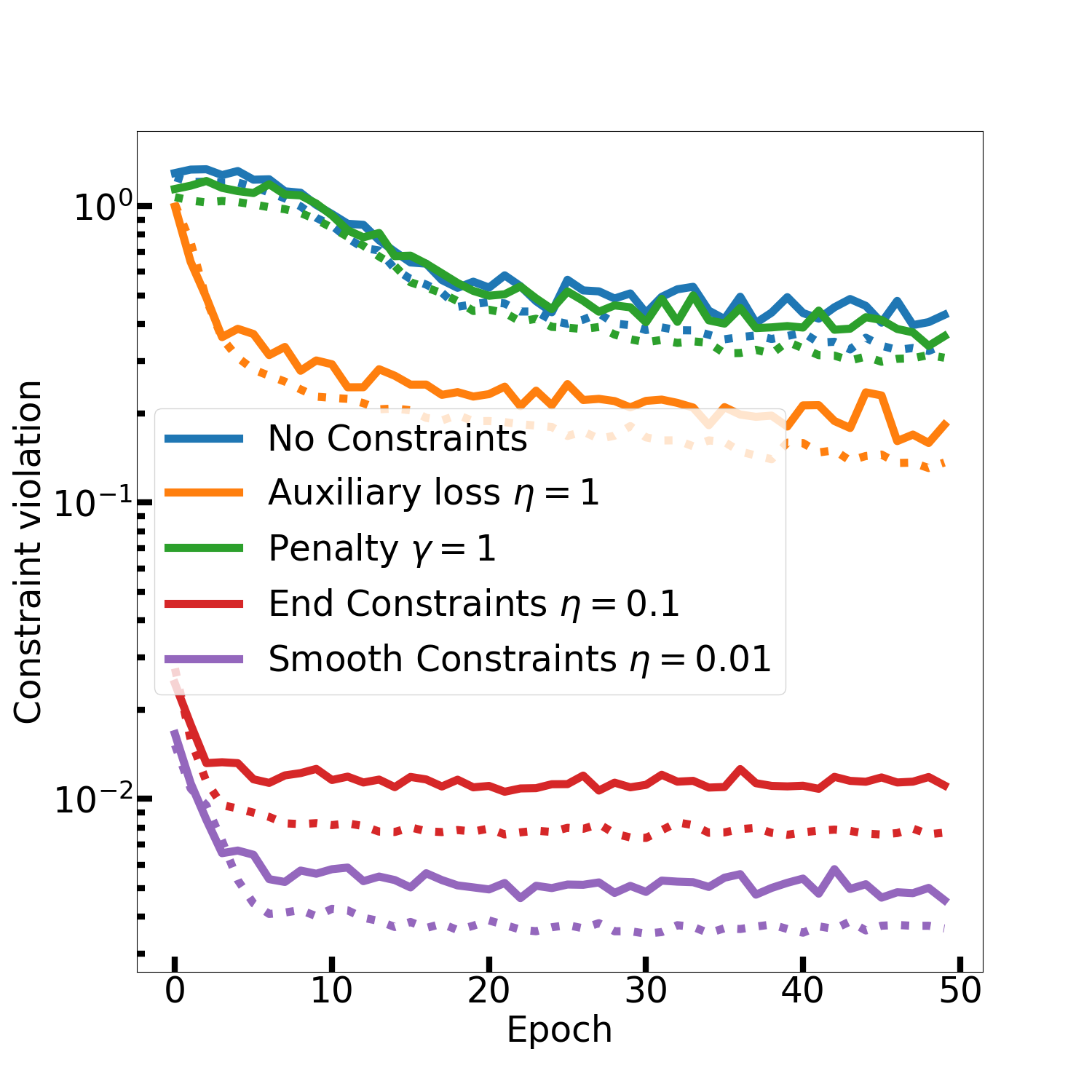}
    \end{center}
\caption{Comparison of the different ways of adding constraints to neural network trained on 100 noisy vector fields. Left: Mean squared loss. Right: Maximum constraint violation. Each run was repeated three times, the solid/dashed lines are the average based on validation/training data.}
\label{fig:imagedenoising}
\end{figure}

Once again, we use the modified versions of smooth-constraints and end-constraints. 
\cref{fig:imagedenoising_example} depicts a field example and the fields predicted by the various methods, while \cref{fig:imagedenoising} depicts training and validation for the various models.
The results on the test set are shown in~\cref{tab:imagedenoising}.

\begin{table}
\begin{center}
\begin{tabular}{ l|r|r }
Constraints & MSE  & CV \\
\hline
No constraints & 0.059 & 2.8 \\
Auxiliary loss $\eta=1$ & 0.069 & 1.0  \\
Penalty $\gamma=1$ (ours) & 0.059 & 2.3 \\
End constraints $\eta=10^{-1}$ (ours) & \textbf{0.051} & 0.14 \\
Smooth constraints $\eta=10^{-2}$ (ours) & 0.059 & \textbf{0.062} \\
\end{tabular}
\end{center}
\caption{Vector field denoising test results. Mean square error (MSE) and mean constraint violation (CV), are shown for all methods. Each experiment was repeated three times and the average of those runs are shown. The best performing model is highlighted.}
\label{tab:imagedenoising}
\end{table}

\subsubsection{Vector field denoising - out of sample distribution}

\begin{figure}
\setlength\tabcolsep{1.5pt}
\begin{tabular}{ccccccc}
  \includegraphics[height=17mm]{./figures/imagedenoising/im0.png} &
  \includegraphics[height=17mm]{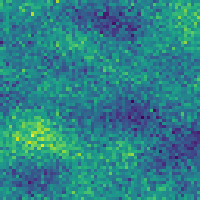} \setlength\tabcolsep{5pt} &
  \includegraphics[height=17mm]{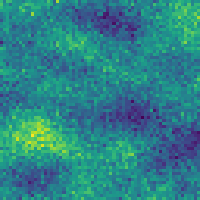} &
  \includegraphics[height=17mm]{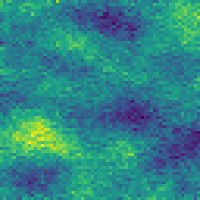} &
  \includegraphics[height=17mm]{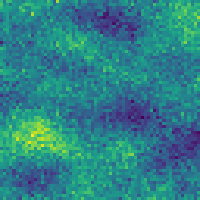} &
  \includegraphics[height=17mm]{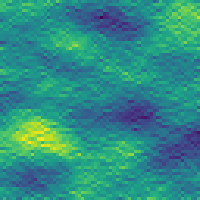} &
  \includegraphics[height=17mm]{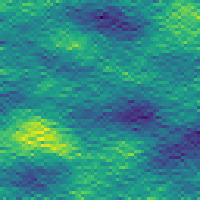}   \\
  \includegraphics[height=17mm]{./figures/imagedenoising/im1.png} &
  \includegraphics[height=17mm]{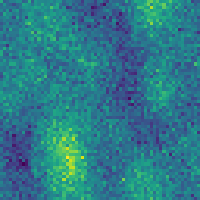} \setlength\tabcolsep{5pt} &
  \includegraphics[height=17mm]{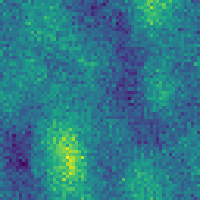} &
  \includegraphics[height=17mm]{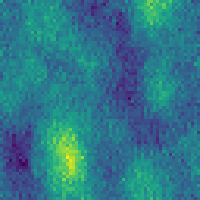} &
  \includegraphics[height=17mm]{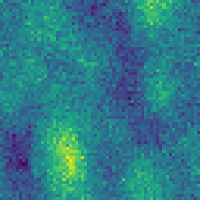} &
  \includegraphics[height=17mm]{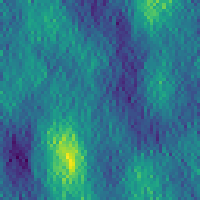} &
  \includegraphics[height=17mm]{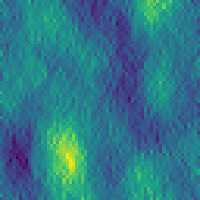} \\
  (a) & (b) & (c) & (d) & (e) & (f) & (g) 
\end{tabular}
\caption{Out of sample vector field denoising example, top row for $u$, bottom row for $v$: (a) true vector field, (b) the noisy input to the neural networks; 
(c)-(g) are the denoised fields from neural networks: (c) no constraints, (d) regularization $\gamma=1$, (e) penalty $\eta=1$, (f) end constraints $\gamma=0.1$, (g) smooth constraints $\gamma=0.01$. 
Models trained on fields with $\sigma=1$ used on a test vector field with $\sigma=10$.}
\label{fig:imagedenoising_example_out_of_sample}
\end{figure}

We wish to investigate how the various models generalize. To test this we use the models that were previously trained on data with noise level $\sigma=1$, and test them on vector fields significantly outside the training distribution. We rerun the test set on all the models, but this time with noise level $\sigma=10$. 
The results from this can be seen in \cref{fig:imagedenoising_example_out_of_sample} and \cref{tab:imagedenoising_out_of_sample}.

\begin{table}
\begin{center}
\begin{tabular}{ l|r|r }
Constraints & MSE & CV \\
\hline
No constraints &  56 & 167 \\
Auxiliary loss $\eta=1$ & 168 & 92 \\
Penalty $\gamma=1$ (ours) &  55 & 165 \\
End constraints $\eta=10^{-1}$ (ours) & 64 & 0.42 \\
Smooth constraints $\eta=10^{-2}$ (ours) & \textbf{33} & \textbf{0.10} \\
\end{tabular}
\end{center}
\caption{Out of sample vector field denoising test results. Mean square error (MSE) and mean constraint violation (CV), are shown for all methods. All the models were trained on fields with $\sigma=1$, but are here being tested on fields with $\sigma=10$, which represent images significantly outside the probability distribution that the model was trained on. 
Each experiment was repeated three times and the average of those runs are shown. The best performing model is highlighted.}
\label{tab:imagedenoising_out_of_sample}
\end{table}


\section{Discussion \& Conclusion}
\label{sec:dc}
In this work we have proposed different ways of incorporating constraint information in a neural network and successfully demonstrated their effectiveness on three separate modelling problems, namely, the multi-body pendulum, where the constraints are exact, the more realistic water molecules simulation, where the constraints are only approximately represented in the data, and finally a divergence-free vector field denoising problem, which shows the broad applicability of our proposed method. We have incorporated these constraints in three fundamentally different neural network architectures, which also demonstrates that the effectiveness of constraints is largely architecture agnostic. 
In all problems, we note a significant improvement when constraint information is incorporated correctly. While the inference effect of applying constraints is noticeable on the training set, it is generally more pronounced on the validation set as seen for instance in the left panel of Figure~\ref{fig:npendulum_final_test}. This suggests that the constraints are not actually ``learned'' during normal training, but rather just memorized, and as such we get considerably larger inference differences between the various methods on the validation data than on the training data. 
In the vector field denoising problem our projection constraints provide a modest improvement when tested on fields with similar noise distribution, but where the constraints really come to effect is when we introduce out-of-sample fields as seen in \cref{fig:imagedenoising_example_out_of_sample} and \cref{tab:imagedenoising_out_of_sample} here the projection constraints ensures that the constraint violation remains low and especially the smooth projection constraint really shines and outperforms all other methods, which is generally what we see happen when the problems get more difficult. 

The boost in prediction precision when adding constraint information is valuable. However, for some fields, such as robotic movements or molecular dynamics, this boost is secondary compared to the decrease in constraint violation, which is far and away the most important aspect. For such fields, a lowering of the constraint violation by several orders of magnitude over standard neural network methods could be the difference between machine learning methods being useless and being the new state-of-the-art. 

Auxiliary regularization variants have  successfully been implemented as evidenced in \cite{stewart2017label,xu2018semantic}, and we similarly find that it can be used to lower constraint violations. But they do so at the cost of decreasing the prediction accuracy of the neural network.
Stabilization methods can lead to boosts in network prediction precision as well as a lowering of constraint violation, but only if the penalty strength $\gamma$ is correctly tuned. If $\gamma$ is not chosen correctly for each specific problem, it can easily lead to a degradation of both prediction and constraint violation performance.
Projection constraints seems to be the superior method, when applied in conjunction with auxiliary regularization, and applying the projections smoothly throughout the neural network seems to generally give better results than applying them only at the end of a neural network. Compared to utilizing no constraints information, the use of smooth constraints significantly reduces prediction errors across all tests, in some cases by more than 50\%, while constraint violations are lowered by several orders of magnitudes. 

In terms of computational resources and the difficulty of implementation, auxiliary regularization is the simplest method. Implementing the penalty method is also easy and only adds a simple Jacobian penalty term, which adds very little in terms of computational performance degradation per iteration. The projection methods are generally the most computationally intensive. End projections can be implemented using standard minimization libraries, and can be applied at the end of a neural network. As such they are relatively easy to implement. 
However, for hard problems where the network prediction is far from honouring the constraint, the end projection can fail to converge in the maximum allowed number of projection steps, as seen in \Cref{tab:pendulum}.
Smooth projections can mitigate this problem by applying the projection in each layer and thus ensuring that the constraint violation remains manageable.
Smooth projection constraints are the most difficult to implement as they require custom minimization methods and need to be implemented directly into the neural network architecture.
While the smooth projections method is computationally more expensive per iteration than any of the other methods it can, if implemented efficiently, actually end up speeding the overall learning, since the same level of prediction precision can often be achieved in a fraction of the number of training epochs, as evidenced by the figures in Appendix~\ref{app:additional_examples}.
In general, smooth projection is the most stable method and produces the best results, but it comes at a price and can be non-trivial to implement. In those cases where the downsides to smooth projections are too dominant end projections offers a much easier implementation and has significantly lower computational demands.

Incorporating constraint information successfully into neural network training requires careful consideration as evidenced by all the above. When implemented wrongly/sub-optimally it can easily lead to worse results than training without constraints. One important aspect in this regard is the number of projection steps which in general needs to be kept as low as possible in order for the neural network to efficiently learn when utilizing backpropagation. 
As such future studies in this field should investigate the effect of more advanced minimization techniques as well as the effect of analytic backward functions for the minimization. Alternatively, evolutionary learning which does not incorporate backpropagation and should thus be able to completely circumvent this problem would also be interesting to investigate. Similarly, we set $\bfH=\bfI$ in the penalty term defined in Section~\ref{sec:stabilization}, but know that more advanced expressions are commonly used in traditional DAEs. 
Another avenue worth exploring is the inclusion of second order constraint information, which can be implemented automatically using the automatic differentiation might be interesting to investigate. Note though that the combination of first and second order constraints generally removes the convex property from the constraints, which means that constraints are more likely to get trapped in local minima.

\section*{Data availability}
The code and data for this project is available at \url{https://github.com/tueboesen/Constrained-Neural-Networks}

\appendix

\section{Network architecture}
\label{app:network}
In this work we used two different neural networks, a 3D rotation-translation equivariant network and a non-equivariant mimetic network \cite{eliasof2021mimetic}. Both networks are residual graph convolutional neural networks, with a learnable stepsize that starts out very small, which ensures that the initial network output is very similar to the best guess (the input).

\subsection{Equivariant network}
Our equivariant network is written using the e3nn software \cite{mario_geiger_2021_4745784}, which allows for any geometric tensor to be accurately represented, and allows meaningful interactions between any such 
objects \cite{thomas2018tensor}. Our network is inspired by \cite{batzner20223}, and reaches comparable results to the ones shown in their work on MD17. The propagation block for our equivariant network is a simple equivariant convolutional filter, a non-linear activation, and a self-interacting tensorproduct. In this work we limit our network to scalars, pseudo-scalars, vectors, and pseudo-vectors. Our equiavariant network has $\approx 2.8$M parameters and 8 layers.

\subsection{Non-equivariant mimetic network}
Our non-equivariant network is a mimetic network, with a propagational block which computes node averages and gradients as well as higher order products by moving node information through their edge connections. The information transfer between nodes and edges is done in a mimetic fashion \cite{lipnikov2014mimetic}. Our mimetic network has $\approx 5.8$M parameters in 8 layers.

\section{Determining penalty strength}
\label{app:gamma}
\begin{figure}[htb!]
	\begin{center}
      \includegraphics[width=0.45\textwidth]{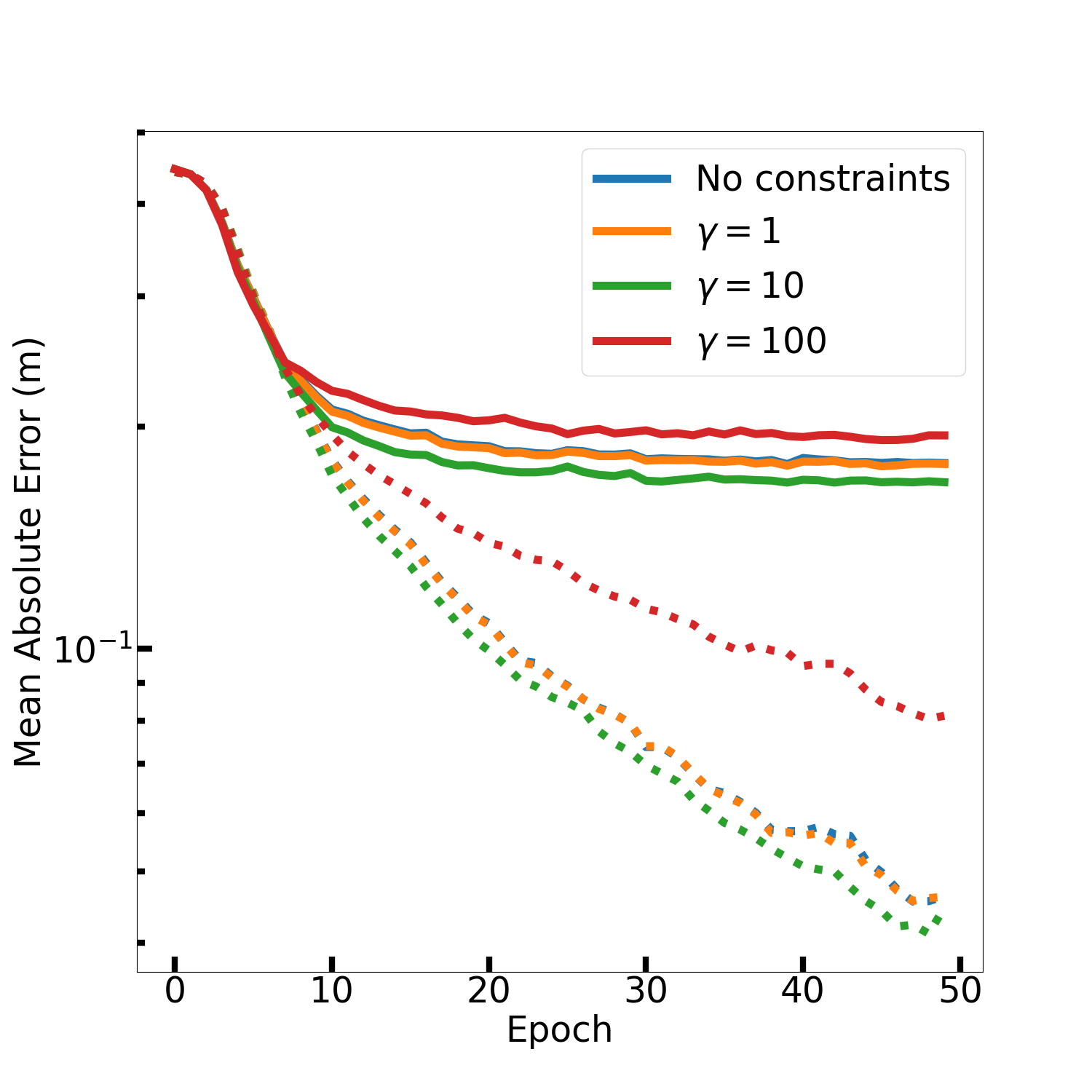}
      \includegraphics[width=0.45\textwidth]{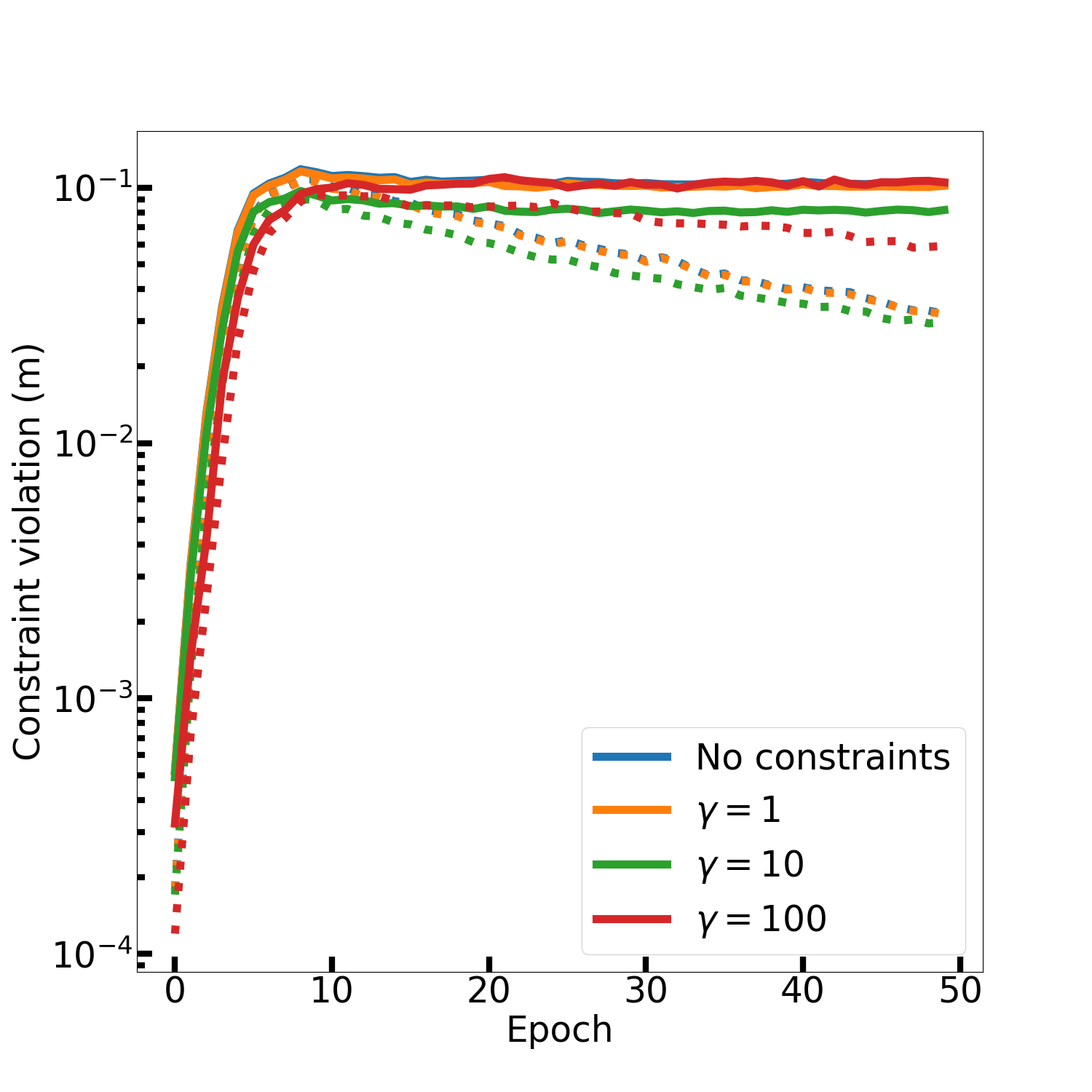}
    \end{center}
\caption{Comparison of neural networks using penalty stabilization with different strengths $\gamma$. All neural networks are trained on 100 samples with $k=100$ on the multi-body pendulum system described in Section~\ref{sec:experiment_pendulum}. Each run has been repeated three times, the solid/dashed lines are the average based on validation/training data. Note that the blue line is mostly hidden beneath the orange line}
\label{fig:npendulum_gamma_sweep_100}
\end{figure}

\begin{figure}[htb!]
	\begin{center}
      \includegraphics[width=0.45\textwidth]{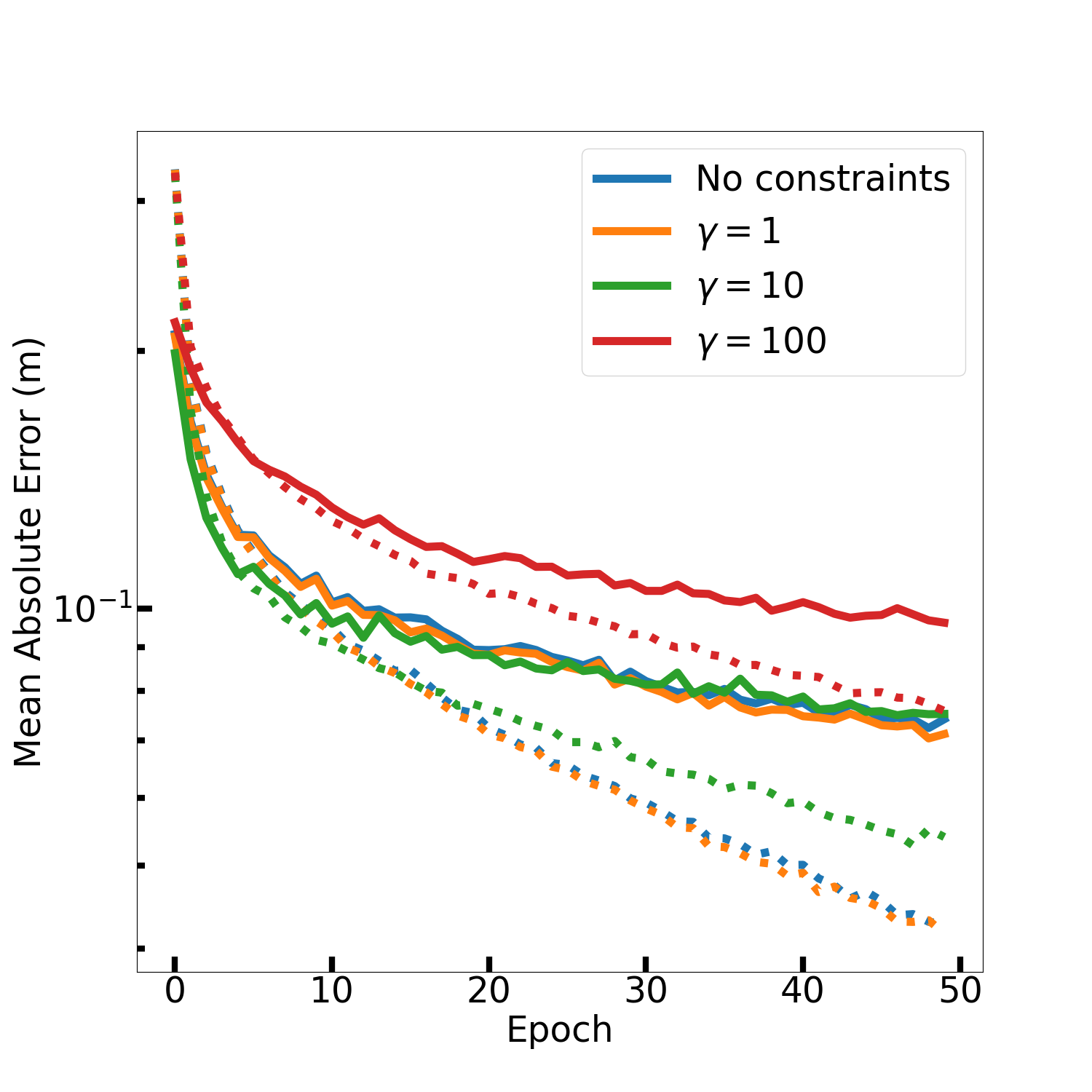}
      \includegraphics[width=0.45\textwidth]{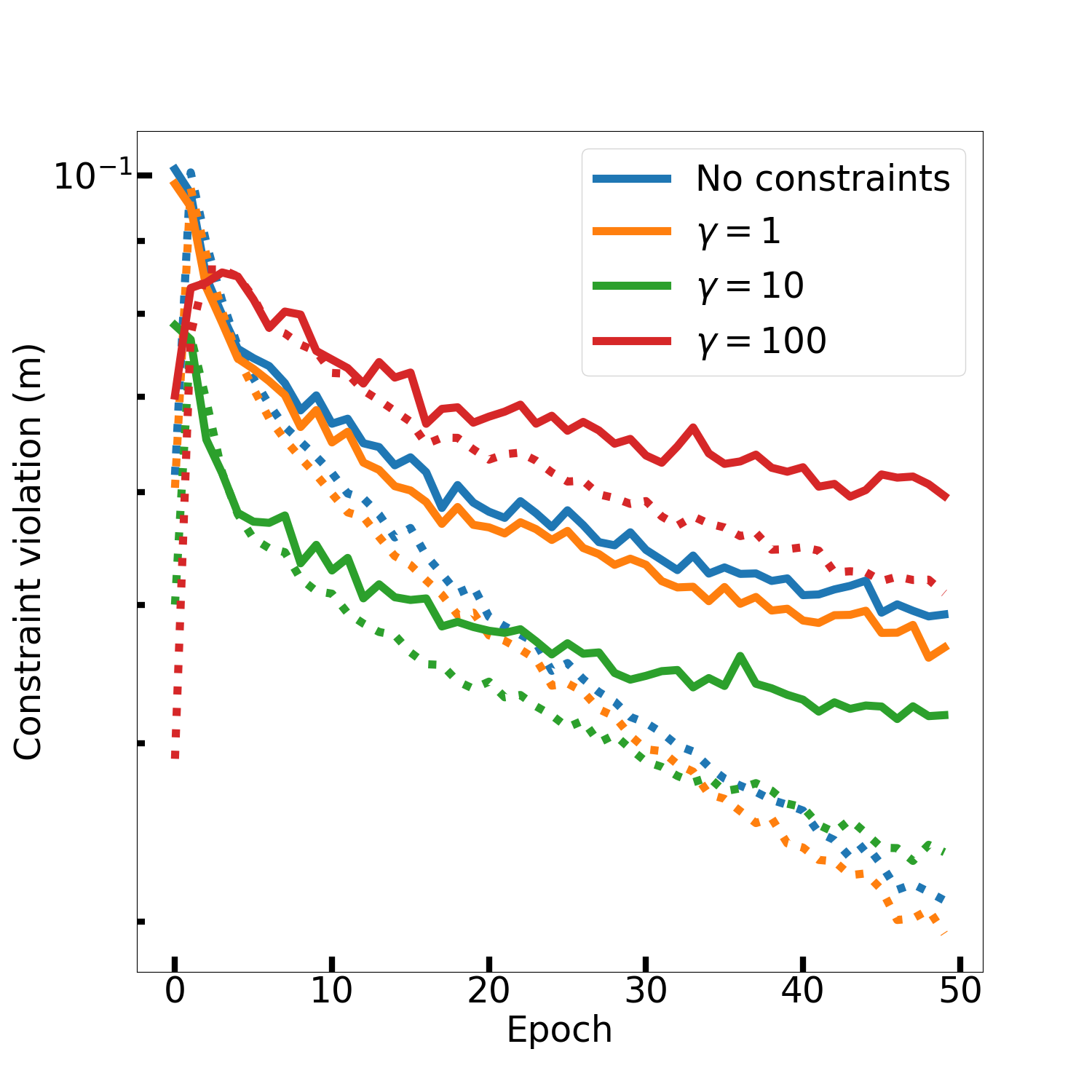}
    \end{center}
\caption{Comparison of neural networks using penalty stabilization with different strengths $\gamma$. All neural networks are trained on 1000 samples with $k=100$ on the five-body pendulum system described in Section~\ref{sec:experiment_pendulum}. Each run has been repeated three times, the solid/dashed lines are the average based on validation/training data. Note that the blue line is mostly hidden beneath the orange line.}
\label{fig:npendulum_gamma_sweep_1000}
\end{figure}

The penalty stabilization method mentioned in Section~\ref{sec:stabilization} contains the hyperparameter $\gamma$, which determines the strength of the penalty. Note that for numerical stability we limit the maximum change of our prediction to 10\% for a single penalty term. Having a maximum allowed change from a single penalty was found to be a crucial necessity since even relative small constraint violation could in rare cases lead to enormous penalty terms when amplified through an RK4 integration. We set $\bfH=\bfI$ and test various values of $\gamma$ in order to find an appropriate value for it. An appropriate penalty strength is typically somewhere in the range of $0.1-100$, and depends on the number of training samples as well as how hard the problem is. Generally, the larger the training dataset is or the harder a problem is, the smaller $\gamma$ should be. 
\Cref{fig:npendulum_gamma_sweep_100,fig:npendulum_gamma_sweep_1000} shows the training/validation for a five-body pendulum with various penalty strengths.  
The $\gamma$ values chosen in Table~\ref{tab:pendulum}-\ref{tab:water} are a compromise between what is appropriate for low and high number of training samples for those problems.

\section{Training with auxiliary regularization}
\label{app:regularization}
\begin{figure}[htb!]
	\begin{center}
      \includegraphics[width=0.45\textwidth]{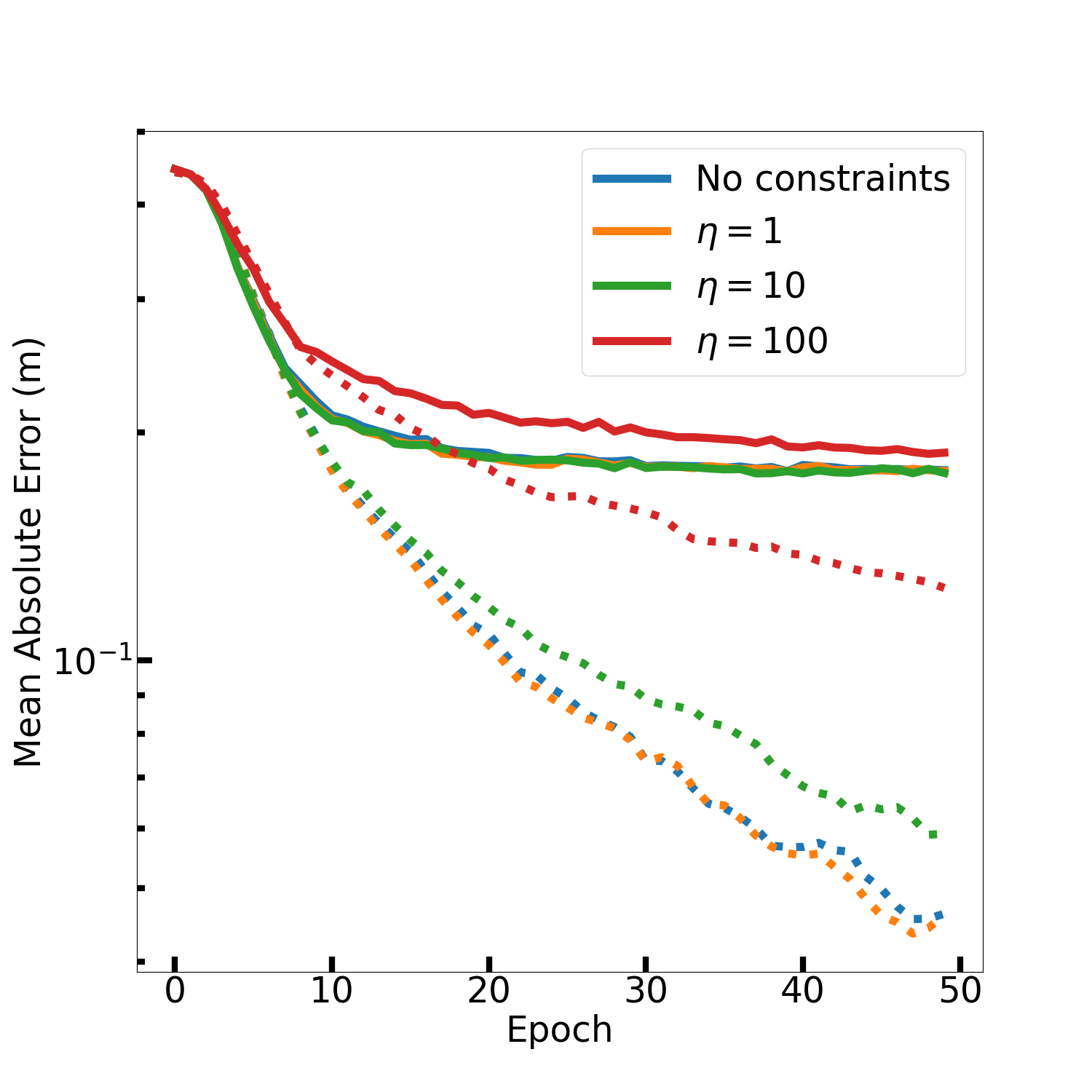}
      \includegraphics[width=0.45\textwidth]{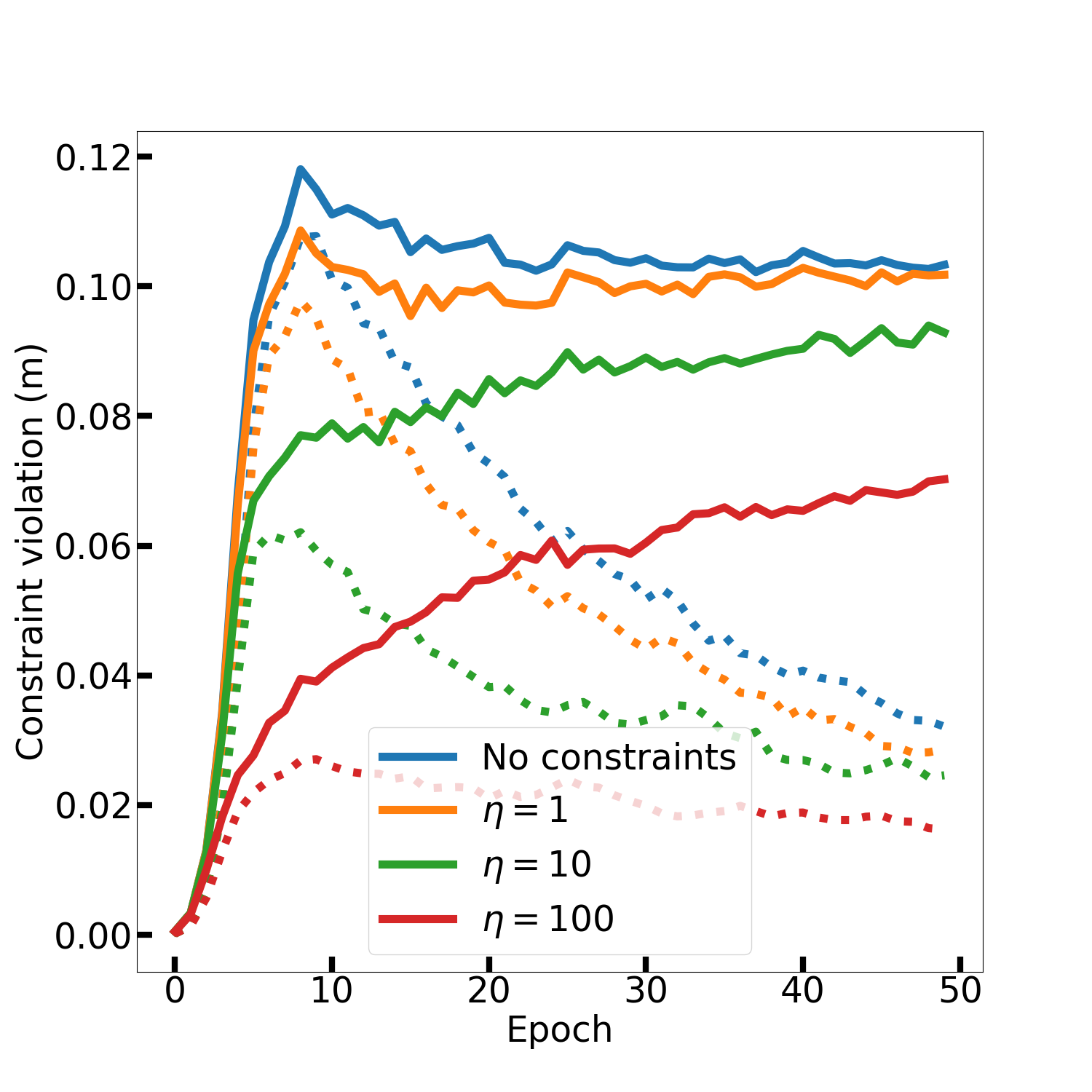}
    \end{center}
\caption{Comparison of neural networks using auxiliary regularization with different strengths $\eta$. All neural networks are trained on 100 samples with $k=100$ on the five-body pendulum system described in Section~\ref{sec:experiment_pendulum}. Each run has been repeated three times, the solid/dashed lines are the average based on validation/training data. Note that the blue line is mostly hidden beneath the orange line.}
\label{fig:npendulum_regularization}
\end{figure}

The stabilization method mentioned in Section~\ref{sec:regularization} employs 
auxiliary regularization, which contains the hyper parameter $\eta$ that determines the strength of the regularization. With auxiliary regularization it is easy to minimize constraint violation, but it does so at the cost of overall learning as seen in Figure~\ref{fig:npendulum_regularization}.

\section{Additional training/validation examples}
\label{app:additional_examples}

Here we include the training and validation information, for all the runs shown in Table~\ref{tab:pendulum} and \ref{tab:water}.

\begin{figure}[htb!]
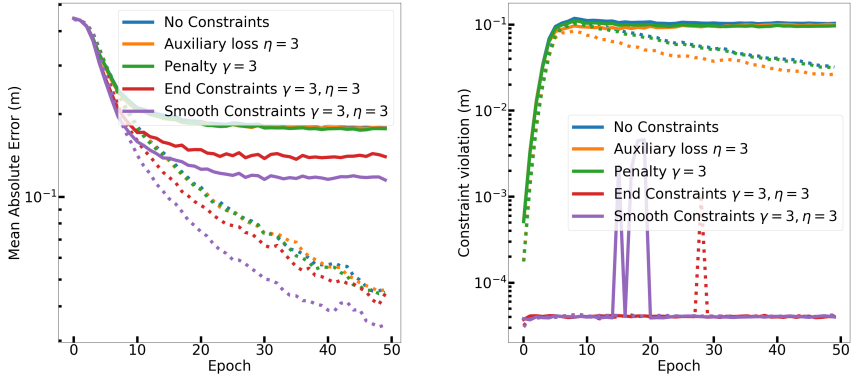

	\begin{center}
      \includegraphics[width=0.45\textwidth]{./figures/comparison/ntrain_100_nskip_100_mae_r.png}
      \includegraphics[width=0.45\textwidth]{./figures/comparison/ntrain_100_nskip_100_cv_mean.png}
    \end{center}
\caption{Comparison of the different ways of adding constraints to neural network trained on 100 multi-body pendulum samples with $k=100$. Left: Mean absolute error. Right: Maximum constraint violation. Each run has been repeated three times, the solid/dashed lines are the average based on validation/training data.}
\label{afig:npendul_ntrain100_k100}
\end{figure}

\begin{figure}[htb!]
	\begin{center}
      \includegraphics[width=0.45\textwidth]{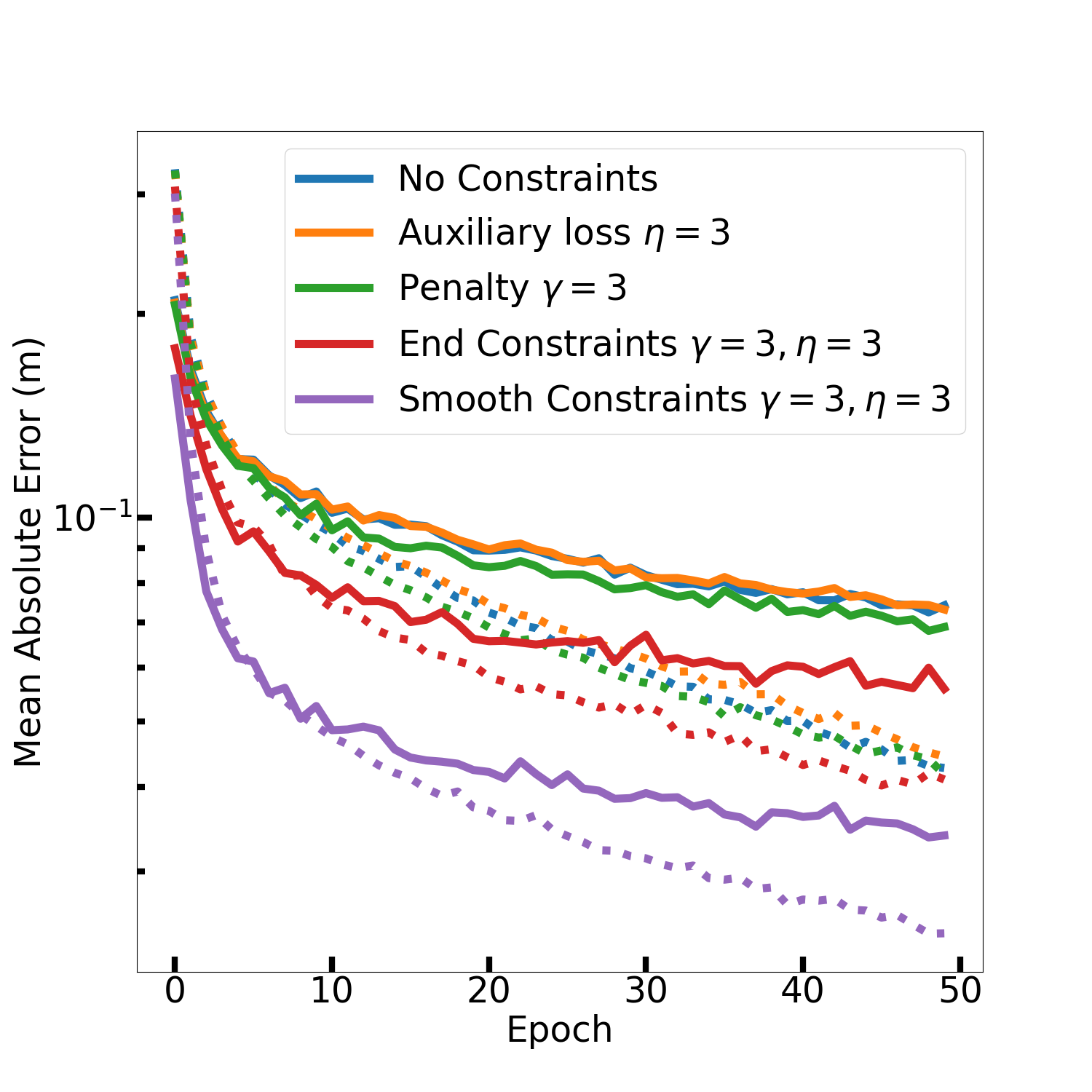}
      \includegraphics[width=0.45\textwidth]{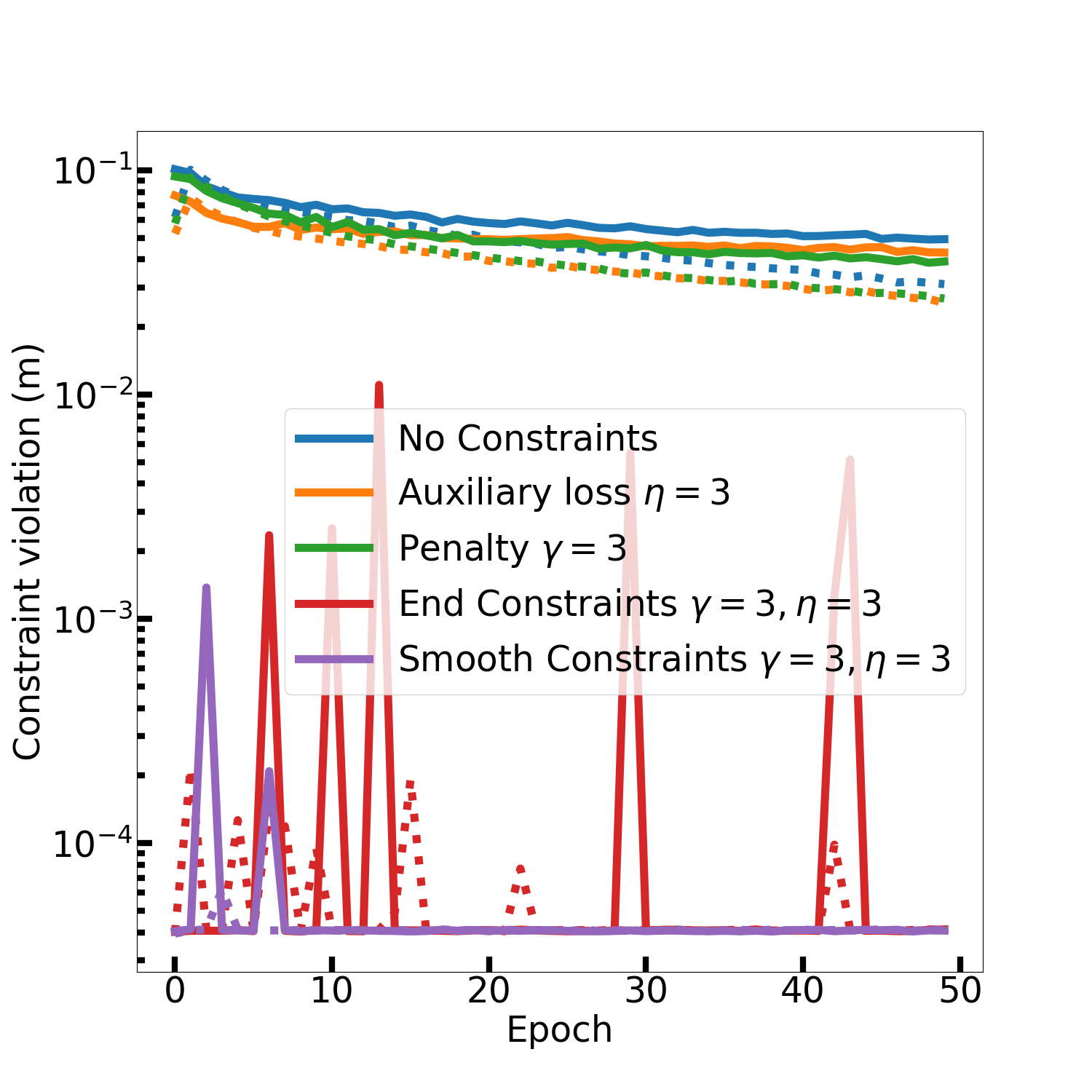}
    \end{center}
\caption{Comparison of the different ways of adding constraints to neural network trained on 1000 multi-body pendulum samples with $k=100$. Left: Mean absolute error. Right: Maximum constraint violation. Each run has been repeated three times, the solid/dashed lines are the average based on validation/training data.}
\label{afig:npendul_ntrain1000_k100}
\end{figure}

\begin{figure}[htb!]
	\begin{center}
      \includegraphics[width=0.45\textwidth]{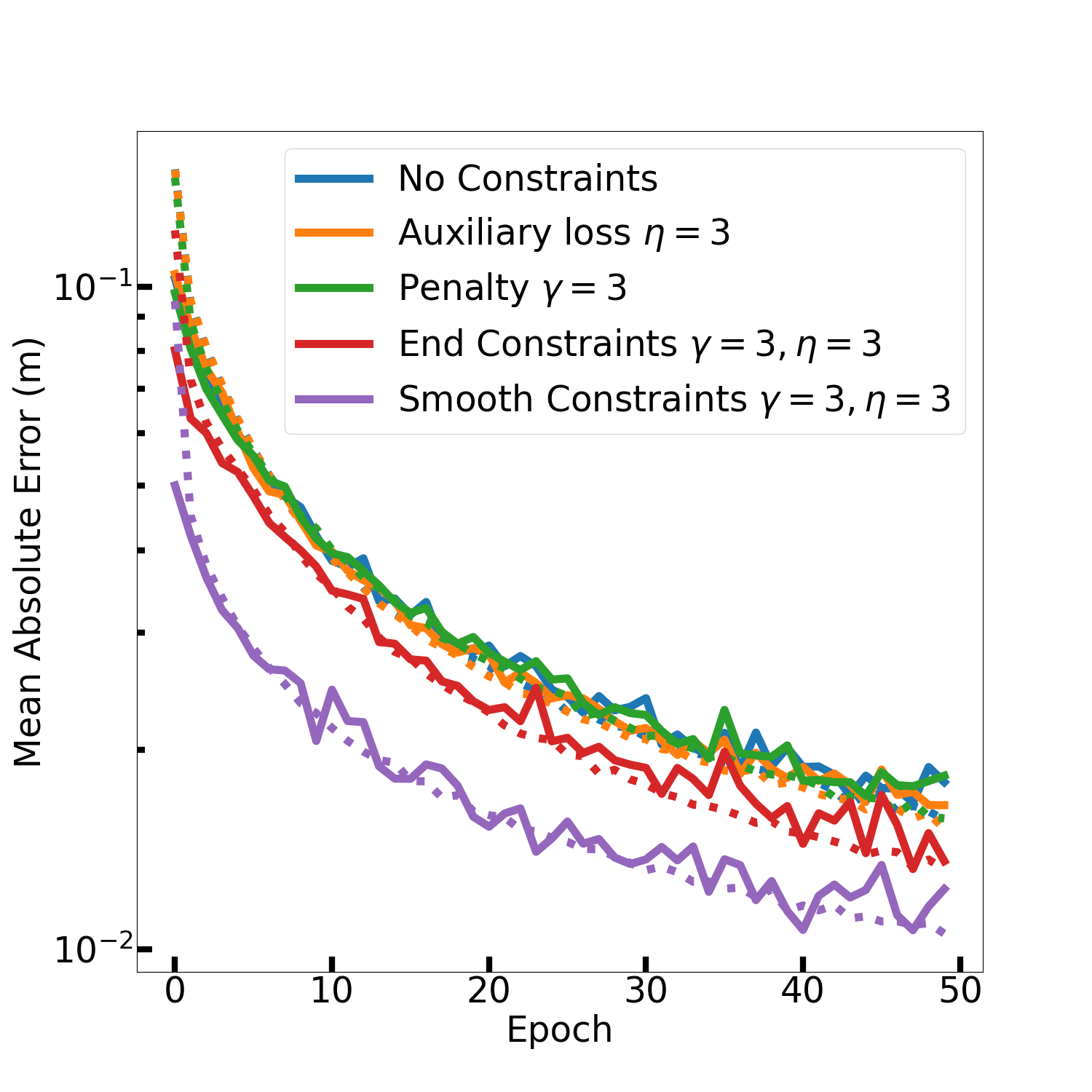}
      \includegraphics[width=0.45\textwidth]{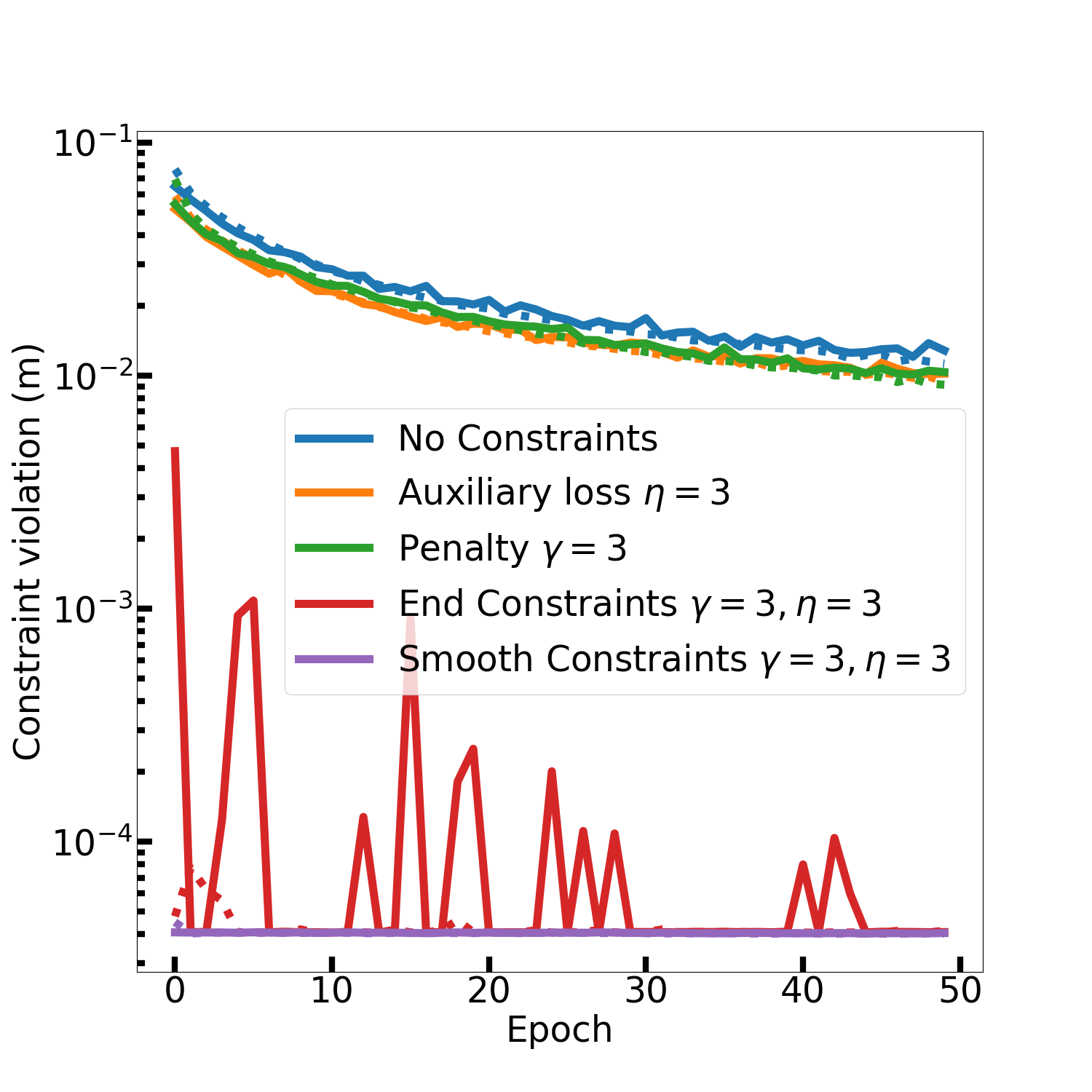}
    \end{center}
\caption{Comparison of the different ways of adding constraints to neural network trained on 10000 multi-body pendulum samples with $k=100$. Left: Mean absolute error. Right: Maximum constraint violation. Each run has been repeated three times, the solid/dashed lines are the average based on validation/training data.}
\label{afig:npendul_ntrain10000_k100}
\end{figure}

\begin{figure}[htb!]
	\begin{center}
      \includegraphics[width=0.45\textwidth]{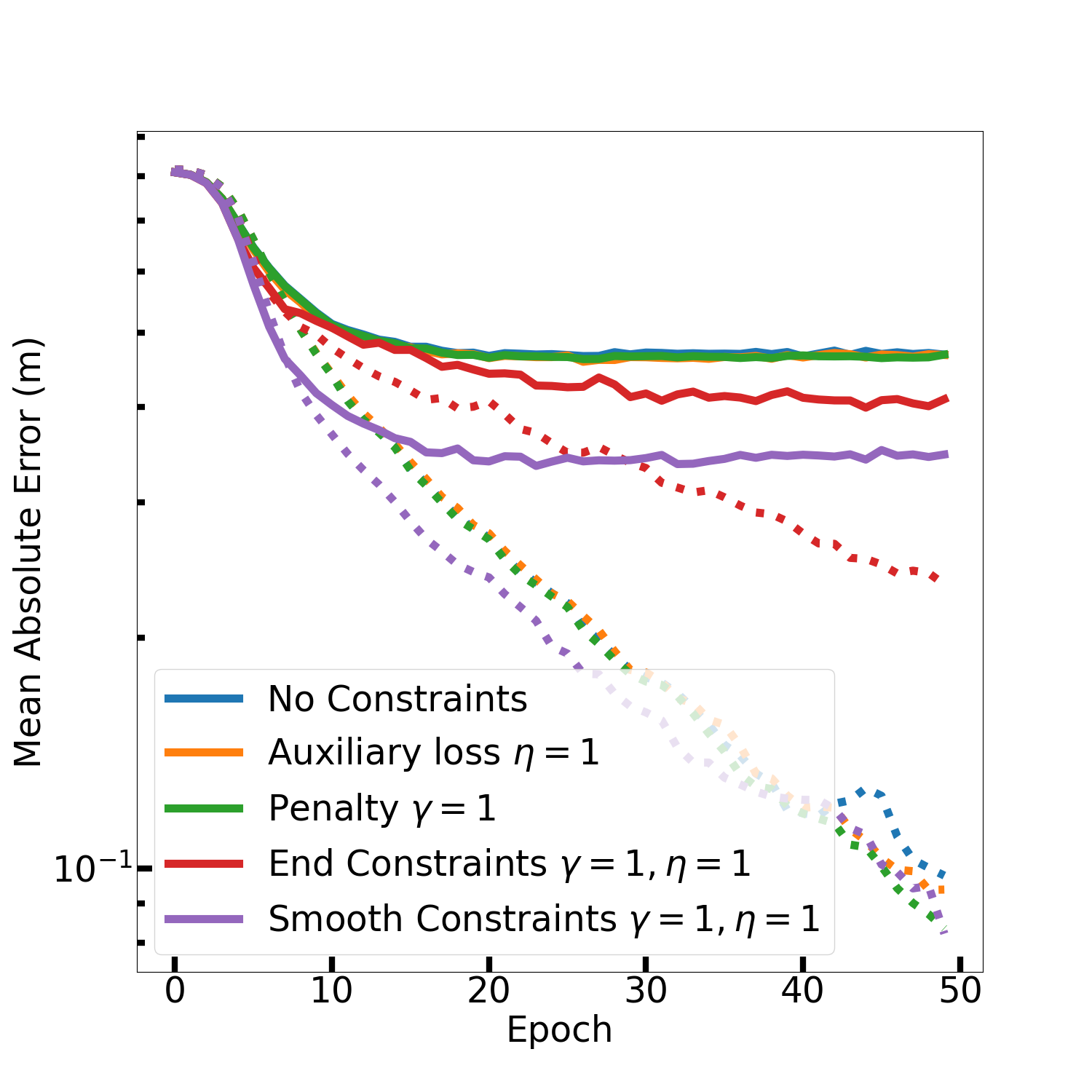}
      \includegraphics[width=0.45\textwidth]{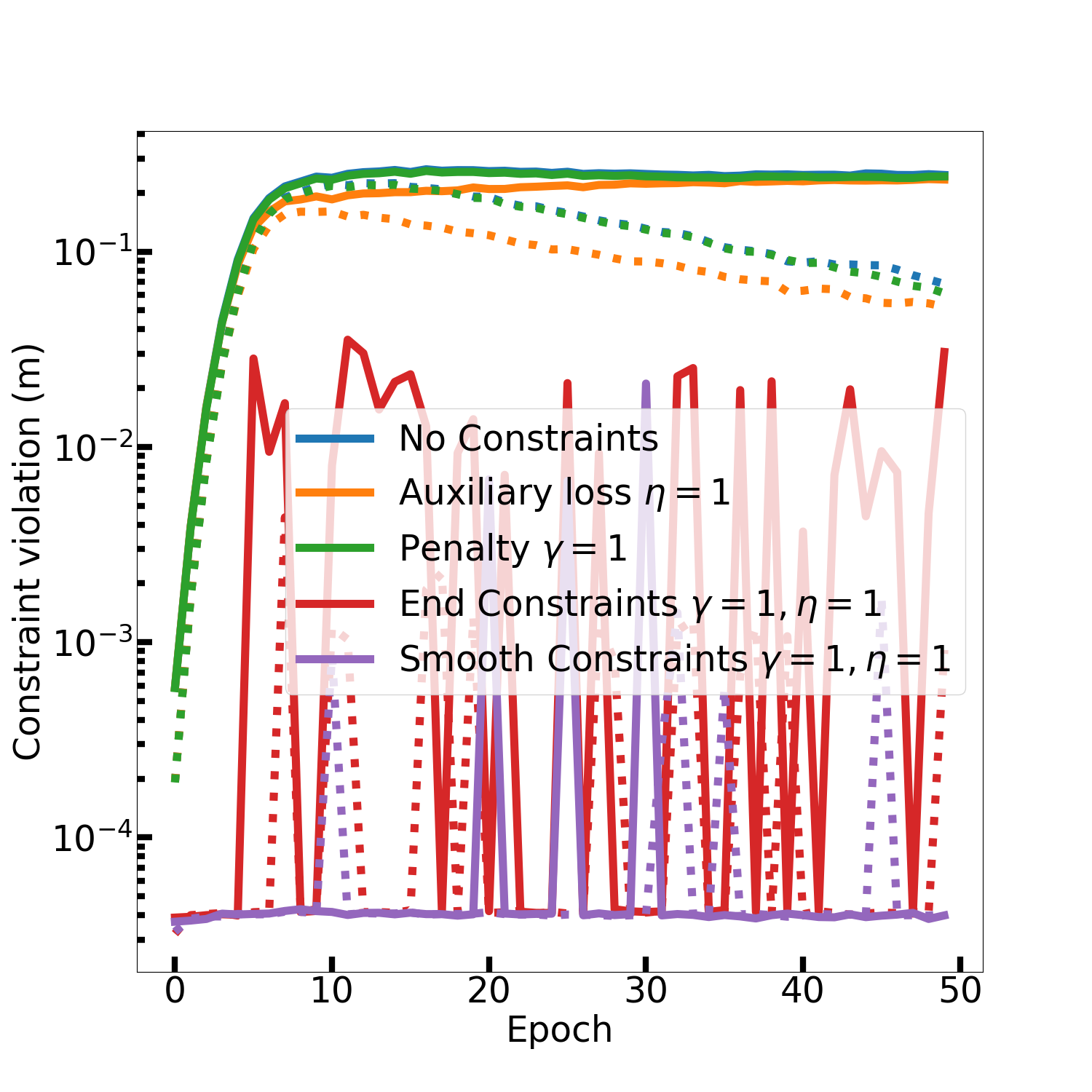}
    \end{center}
\caption{Comparison of the different ways of adding constraints to neural network trained on 100 multi-body pendulum samples with $k=200$. Left: Mean absolute error. Right: Maximum constraint violation. Each run has been repeated three times, the solid/dashed lines are the average based on validation/training data.}
\label{afig:npendul_ntrain100_k200}
\end{figure}

\begin{figure}[htb!]
	\begin{center}
      \includegraphics[width=0.45\textwidth]{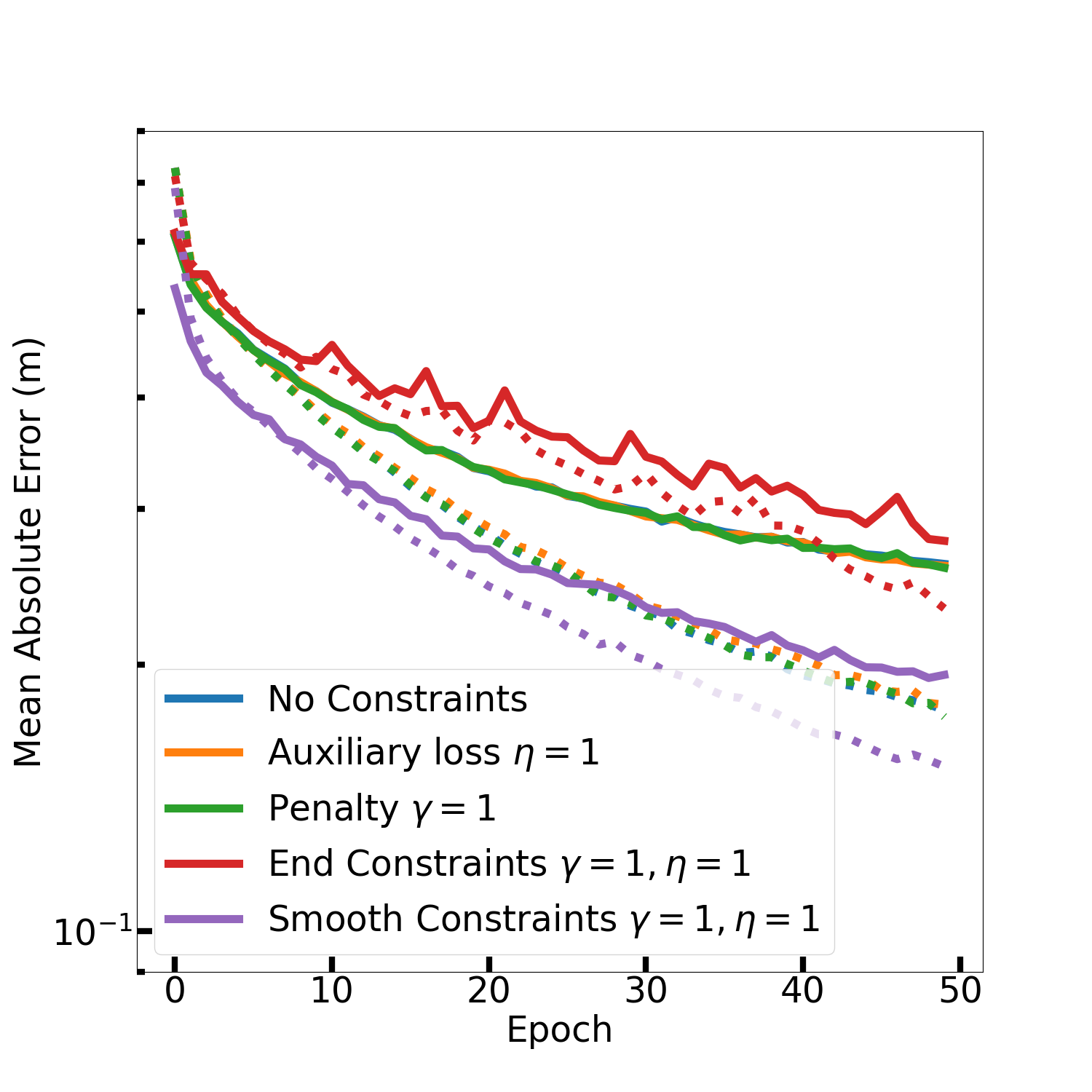}
      \includegraphics[width=0.45\textwidth]{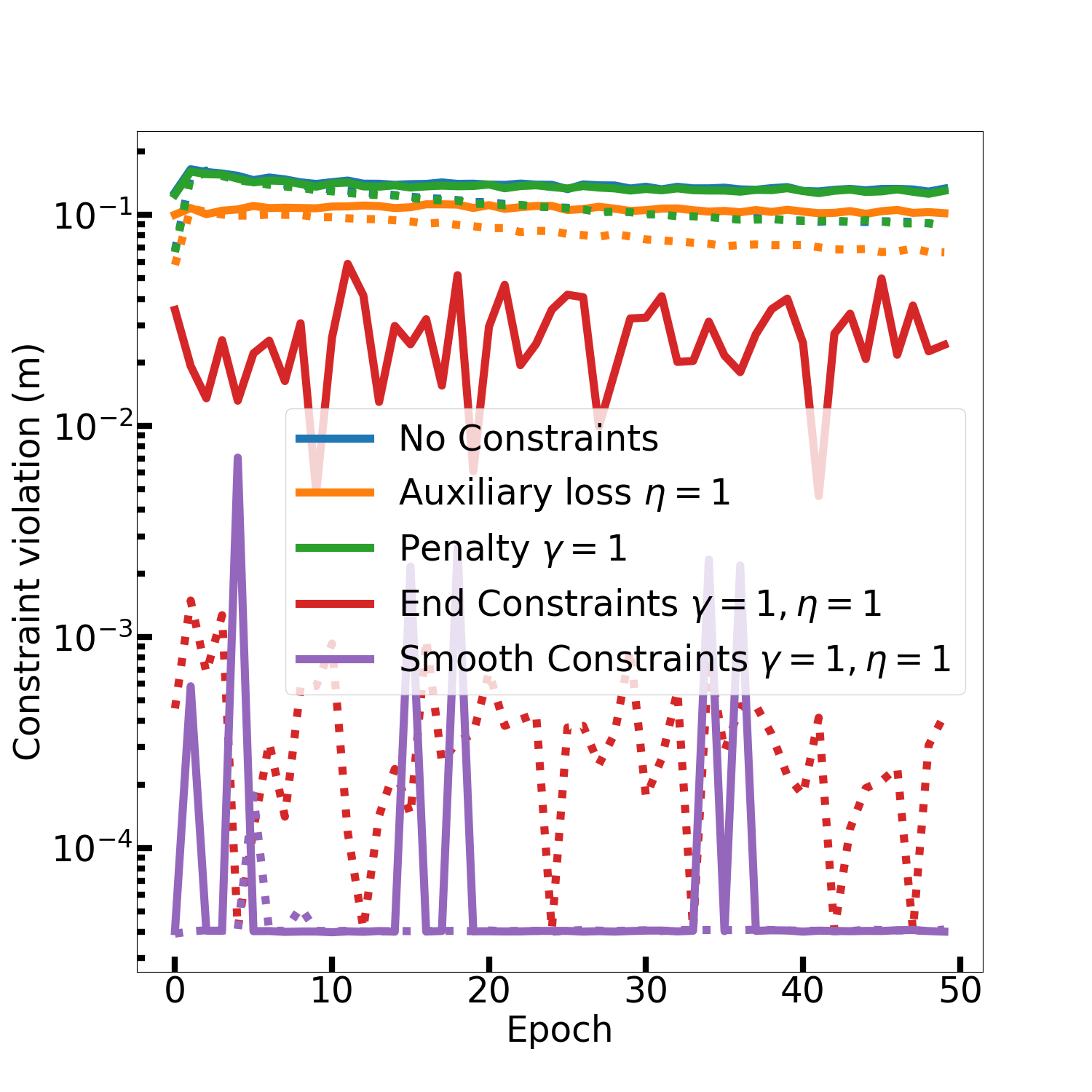}
    \end{center}
\caption{Comparison of the different ways of adding constraints to neural network trained on 1000 multi-body pendulum samples with $k=200$. Left: Mean absolute error. Right: Maximum constraint violation. Each run has been repeated three times, the solid/dashed lines are the average based on validation/training data.}
\label{afig:npendul_ntrain1000_k200}
\end{figure}

\begin{figure}[htb!]
	\begin{center}
      \includegraphics[width=0.45\textwidth]{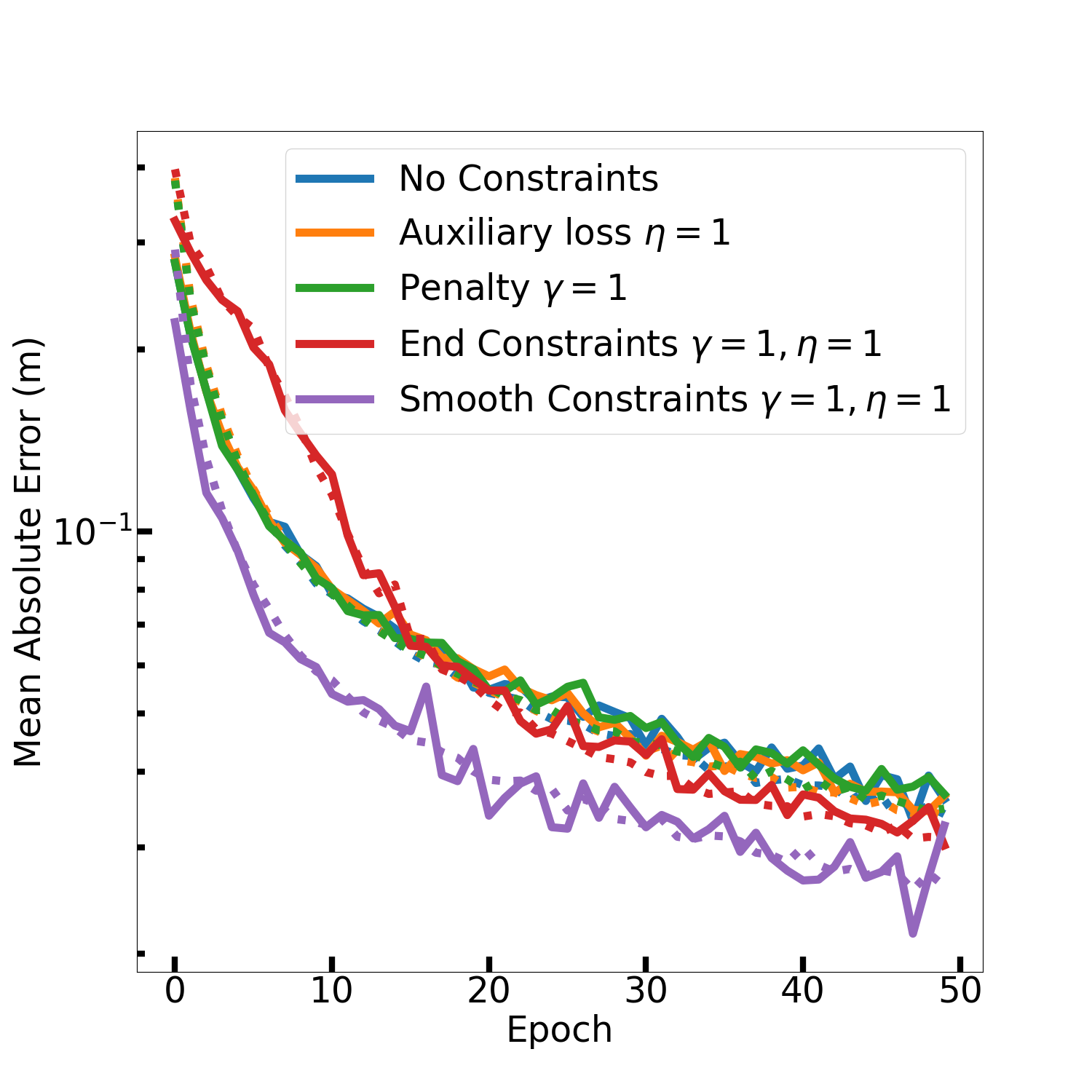}
      \includegraphics[width=0.45\textwidth]{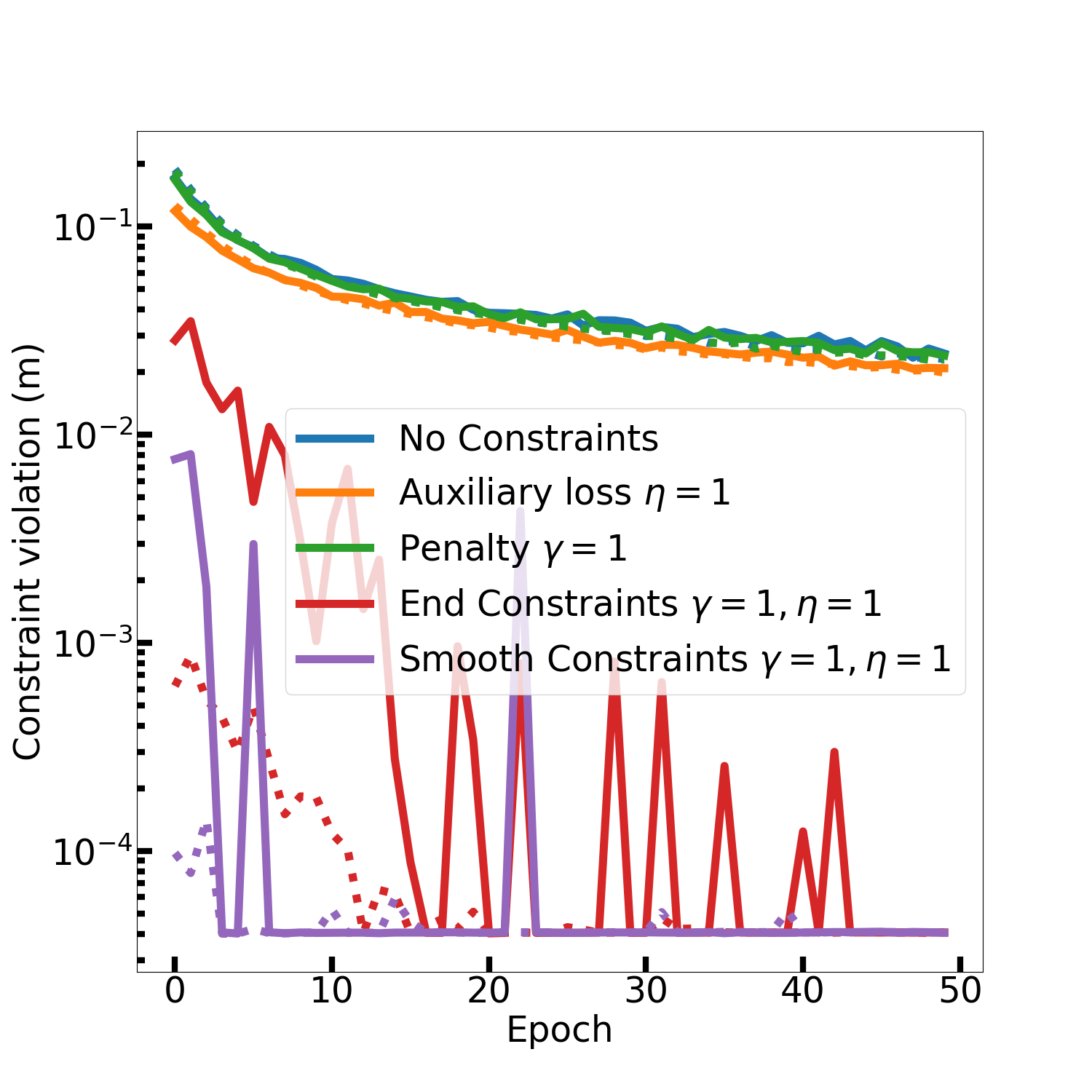}
    \end{center}
\caption{Comparison of the different ways of adding constraints to neural network trained on 10000 multi-body pendulum samples with $k=200$. Left: Mean absolute error. Right: Maximum constraint violation. Each run has been repeated three times, the solid/dashed lines are the average based on validation/training data.}
\label{afig:npendul_ntrain10000_k200}
\end{figure}

\begin{figure}[htb!]
	\begin{center}
      \includegraphics[width=0.45\textwidth]{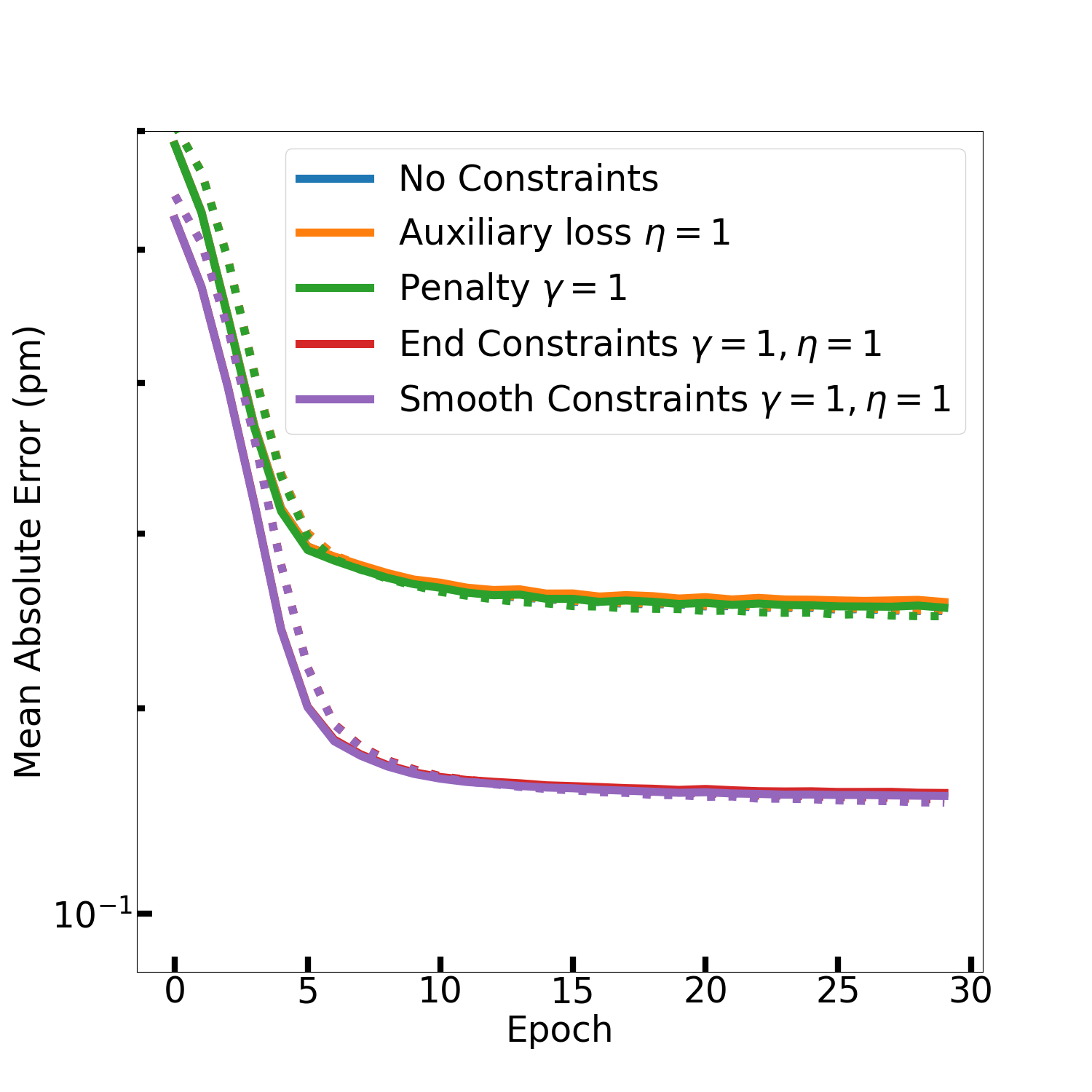}
      \includegraphics[width=0.45\textwidth]{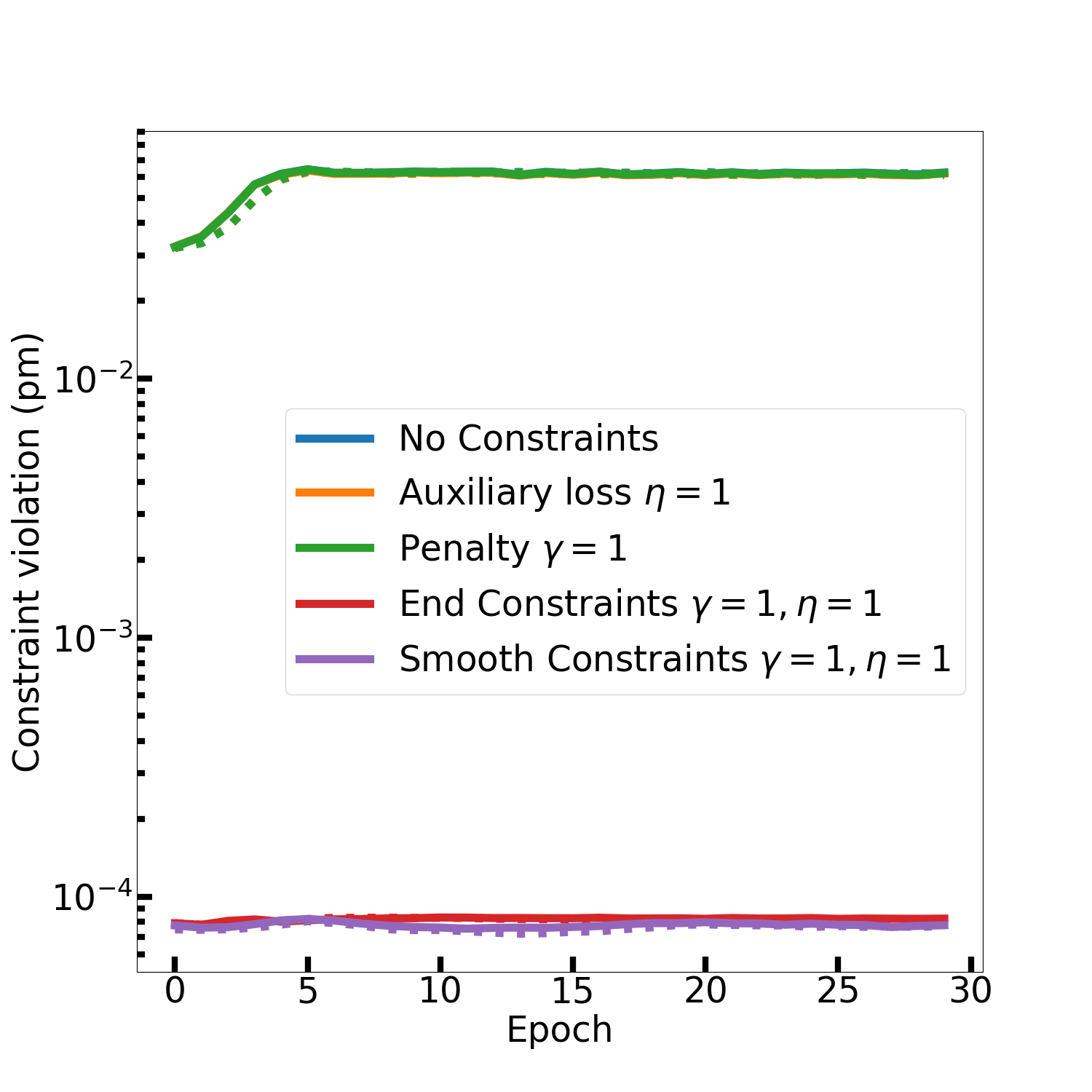}
    \end{center}
\caption{Comparison of the different ways of adding constraints to neural network trained on 100 water molecule samples with $k=50$. Left: Mean absolute error. Right: Maximum constraint violation. Each run has been repeated three times, the solid/dashed lines are the average based on validation/training data.}
\label{afig:water_100}
\end{figure}

\begin{figure}[htb!]
	\begin{center}
      \includegraphics[width=0.45\textwidth]{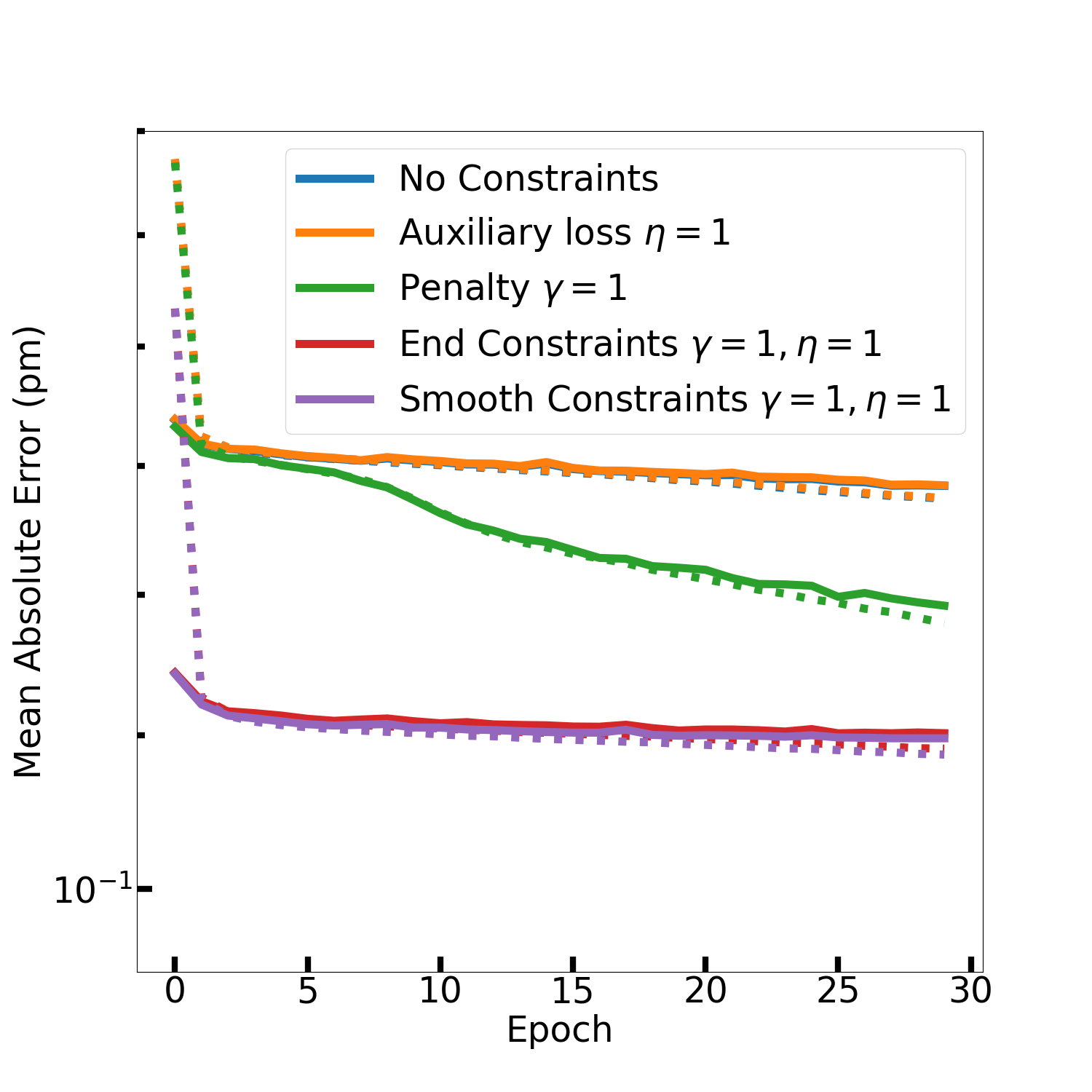}
      \includegraphics[width=0.45\textwidth]{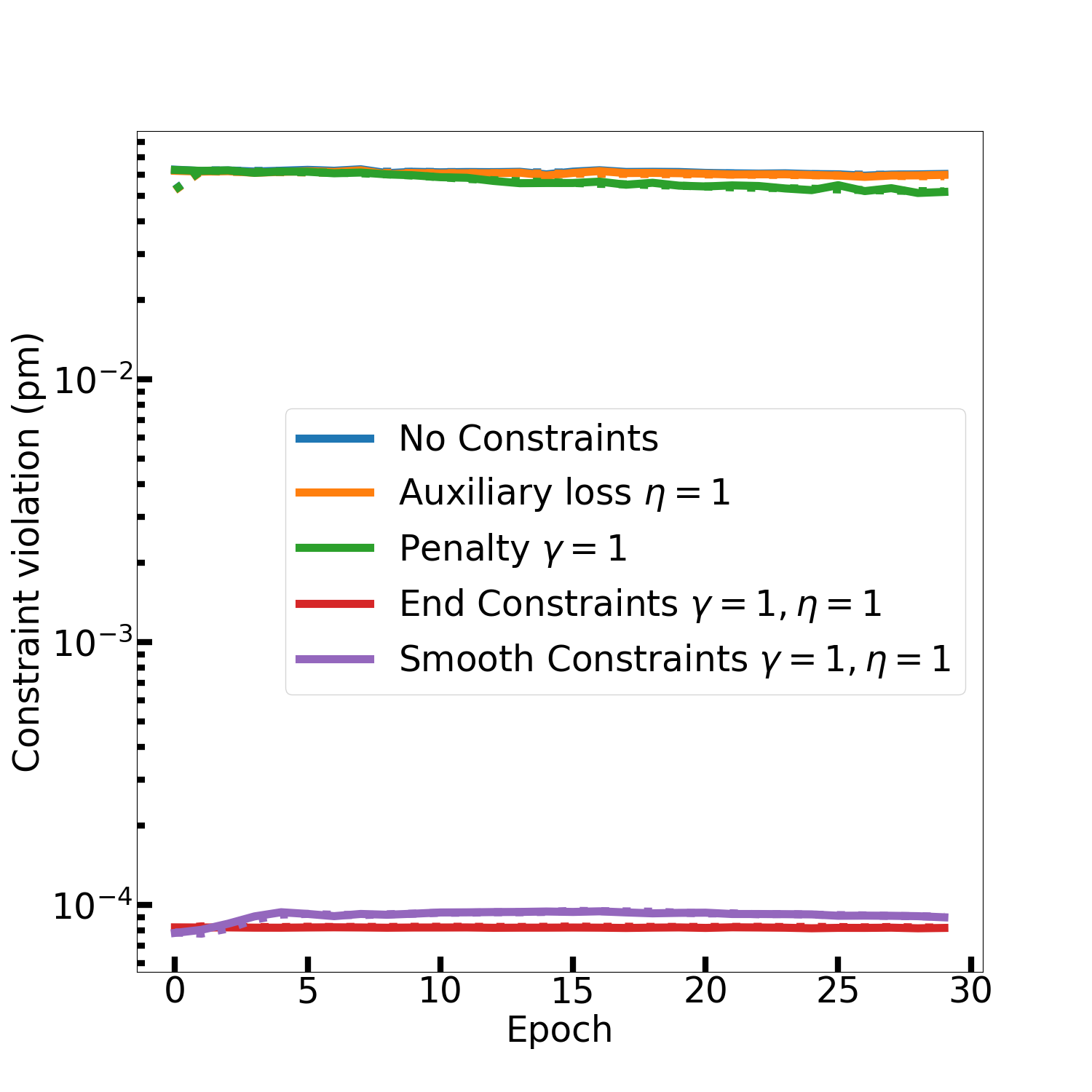}
    \end{center}
\caption{Comparison of the different ways of adding constraints to neural network trained on 1000 water molecule samples with $k=50$. Left: Mean absolute error. Right: Maximum constraint violation. Each run has been repeated three times, the solid/dashed lines are the average based on validation/training data.}
\label{afig:water_1000}
\end{figure}

\begin{figure}[htb!]
	\begin{center}
      \includegraphics[width=0.45\textwidth]{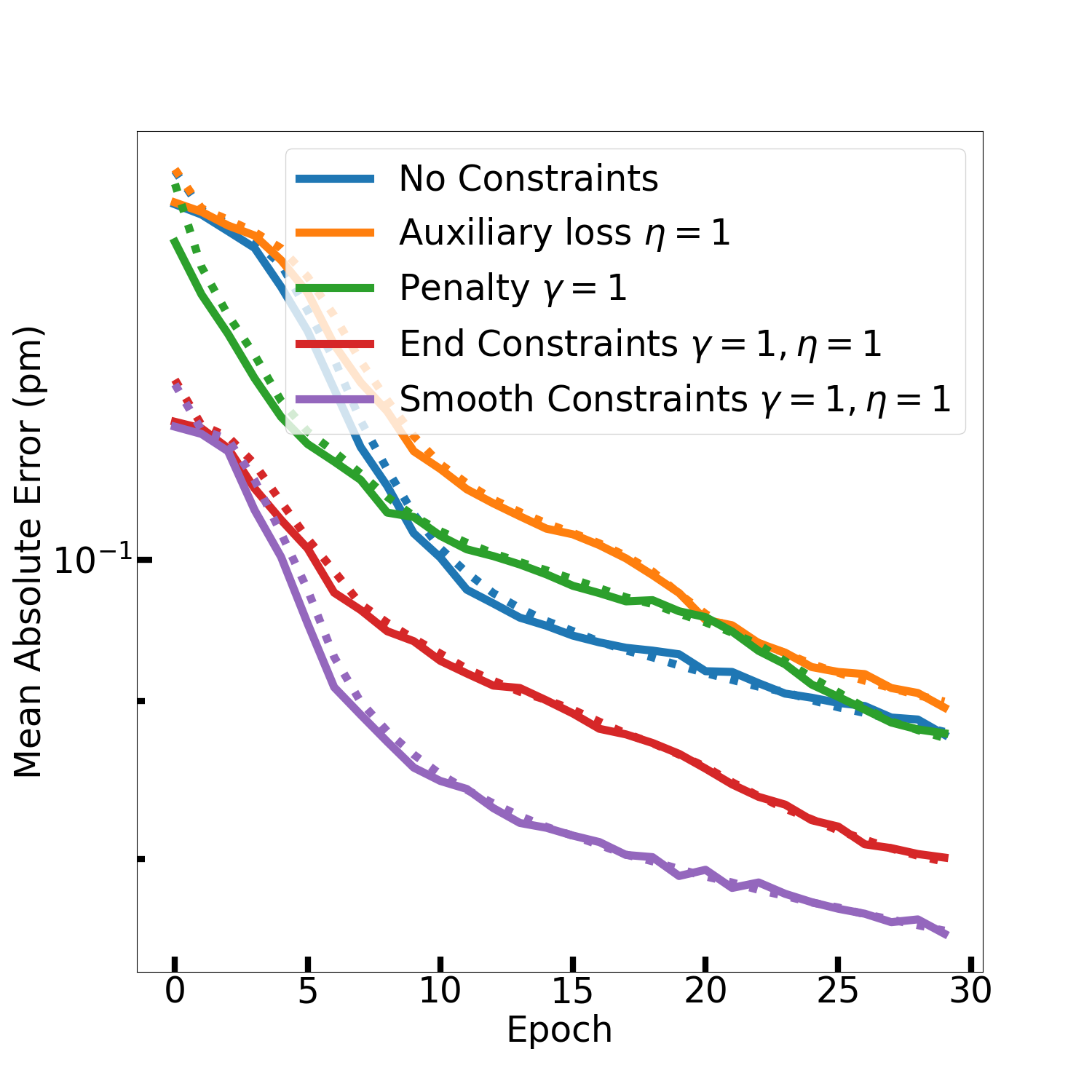}
      \includegraphics[width=0.45\textwidth]{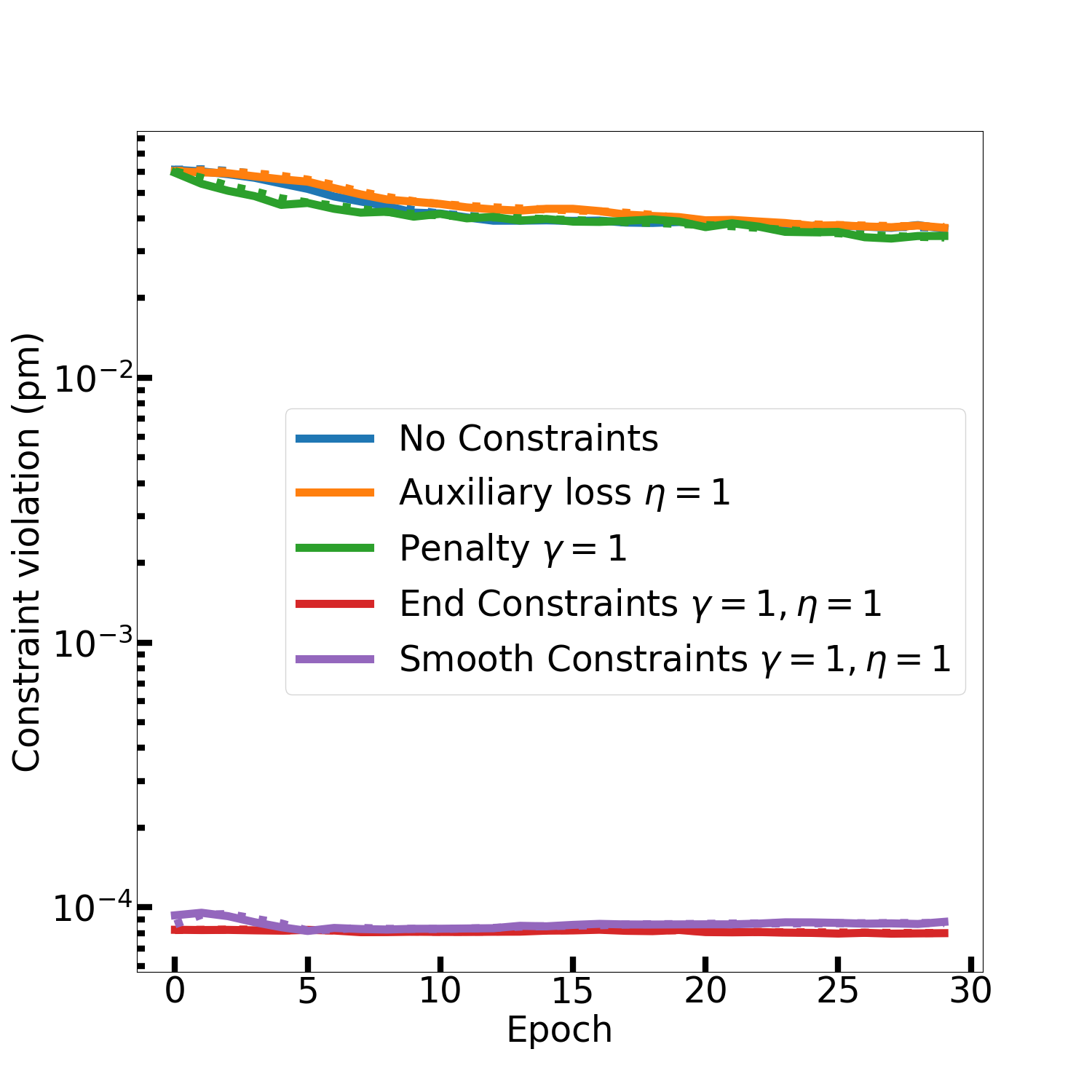}
    \end{center}
\caption{Comparison of the different ways of adding constraints to neural network trained on 10000 water molecule samples with $k=50$. Left: Mean absolute error. Right: Maximum constraint violation. Each run has been repeated three times, the solid/dashed lines are the average based on validation/training data.}
\label{afig:water_10000}
\end{figure}

\bibliographystyle{siamplain}
\bibliography{references}

\end{document}